\pdfoutput=1

\documentclass[11pt]{article}

\usepackage[final]{acl}

\usepackage{times}
\usepackage{latexsym}
\usepackage{amsmath}
\usepackage{amssymb}

\usepackage[T1]{fontenc}

\usepackage[utf8]{inputenc}

\usepackage{microtype}

\usepackage{inconsolata}

\usepackage{graphicx}
\usepackage{subcaption}
\usepackage{booktabs}
\usepackage{xcolor}
\usepackage[most]{tcolorbox}
\usepackage{arydshln}
\usepackage[ruled]{algorithm2e}
\SetKwProg{Fn}{Function}{:}{}
\SetKwProg{Proc}{Procedure}{:}{}
\usepackage{colortbl}
\definecolor{mygray}{gray}{0.8}
\definecolor{yellow_light}{RGB}{255, 235, 140}    
\definecolor{yellow_medium}{RGB}{255, 204, 102}   
\definecolor{yellow_dark}{RGB}{255, 153, 51}      
\definecolor{blue_light}{RGB}{173, 216, 230}    
\definecolor{blue_medium}{RGB}{100, 149, 237}   
\definecolor{blue_dark}{RGB}{90, 150, 210}      
\definecolor{green_light}{RGB}{144, 238, 144}      
\definecolor{green_medium}{RGB}{60, 179, 113}      
\definecolor{green_dark}{RGB}{0, 100, 0}           

\definecolor{blockblue}{RGB}{214,234,248}
\definecolor{blockgreen}{RGB}{221,237,211}
\definecolor{blockorange}{RGB}{254,235,201}
\definecolor{blockpink}{RGB}{251,228,236}

\definecolor{pastelblue}{RGB}{173,216,230}
\definecolor{pastelpink}{RGB}{255,209,220}
\definecolor{pastelgreen}{RGB}{189,230,189}
\definecolor{pastelyellow}{RGB}{255,247,175}
\definecolor{pastelorange}{RGB}{255,218,185}
\tcbuselibrary{skins, breakable}

\definecolor{seg1}{RGB}{242,233,230} 
\definecolor{seg2}{RGB}{232,242,235} 
\definecolor{seg3}{RGB}{235,238,242} 

\newtheorem{theorem}{Theorem}

\newtheorem{proof}{Proof}

\usepackage{extarrows}

\title{\includegraphics[width=0.05\textwidth]{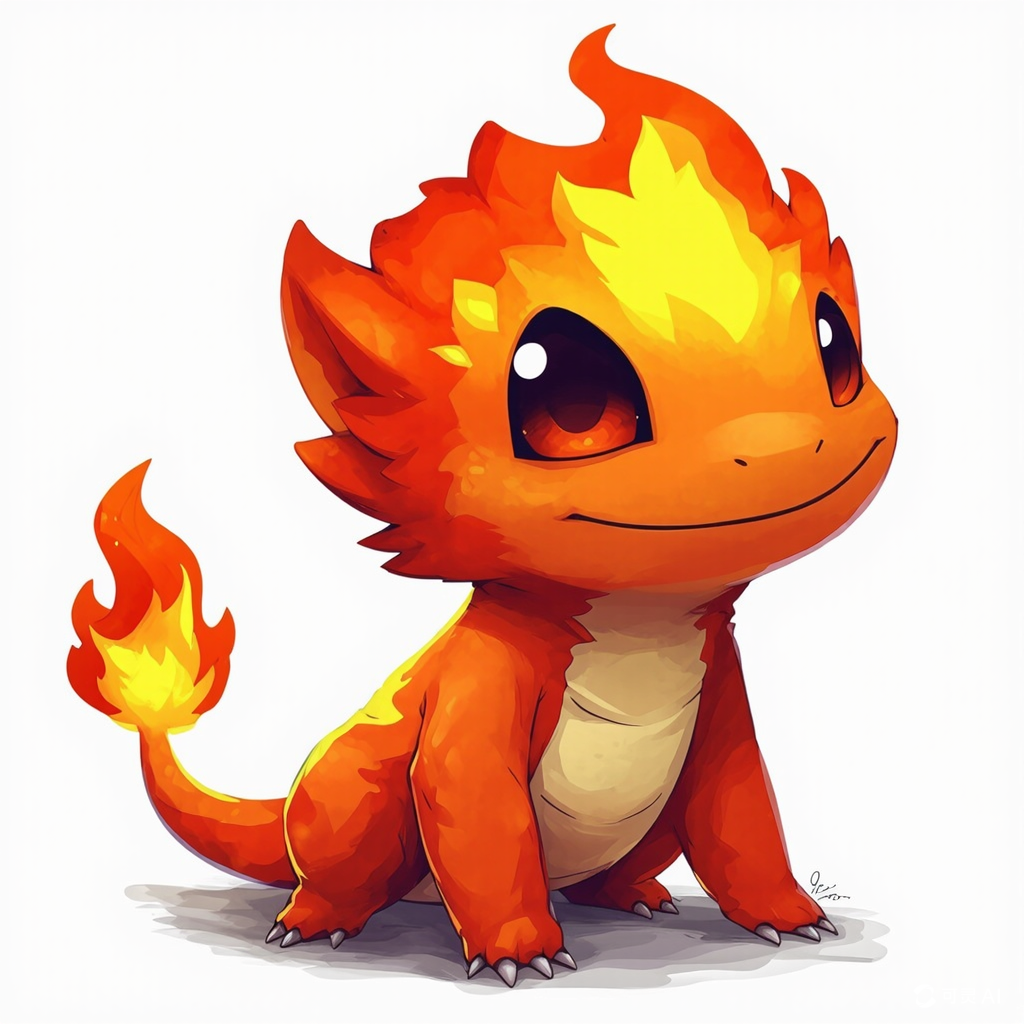}{SalaMAnder}: Shapley-based Mathematical Expression Attribution and Metric for Chain-of-Thought Reasoning}

\author{
 \textbf{Yue Xin\textsuperscript{1,2}}*,
 \textbf{Chen Shen\textsuperscript{2$\dagger\ddagger$}},
 \textbf{Shaotian Yan\textsuperscript{2}},
 \textbf{Xiaosong Yuan\textsuperscript{2}},\\
 \textbf{Yaoming Wang\textsuperscript{1}},
 \textbf{Xiaofeng Zhang\textsuperscript{1,2}}*,
 \textbf{Chenxi Huang\textsuperscript{2}},
 \textbf{Jieping Ye\textsuperscript{2}}
\\ \\
 \textsuperscript{1}Shanghai Jiao Tong University,
 \textsuperscript{2}Alibaba Cloud Computing\\
}

\begin{document}
\maketitle

\def\thefootnote{*}\footnotetext{Work done during an internship at Alibaba Cloud Computing.}
\def\thefootnote{$\dagger$}\footnotetext{Corresponding: Chen Shen<jason.sc@alibaba-inc.com>}
\def\thefootnote{$\ddagger$}\footnotetext{Project Lead.}

\begin{abstract}
Chain-of-Thought (CoT) prompting enhances the math reasoning capability of large language models (LLMs) to a large margin. However, the mechanism underlying such improvements remains unexplored. In this paper, we present \textbf{SalaMAnder} (\textbf{S}h\textbf{a}p\textbf{l}ey-b\textbf{a}sed \textbf{M}athematical Expression \textbf{A}ttribution a\textbf{nd} M\textbf{e}t\textbf{r}ic), a theoretically grounded methodology as well as a mathematically rigorous evaluation metric for quantifying component-level contributions in few-shot CoT reasoning. Concretely, we leverage the Shapley value for mathematical expression attribution and develop an efficient stratified sampling algorithm that significantly reduces the computational complexity. Besides, we develop the \textbf{CoSP} (\textbf{C}ardinality \textbf{o}f \textbf{S}hapley \textbf{P}ositives) metric through covariance analysis. Comprehensive validation across popular LLM models and diverse mathematical benchmarks demonstrates that the CoSP metric within our SalaMAnder framework exhibits a robust monotonic correlation with model performance, not only providing theoretical explanations for the empirical success of existing few-shot CoT but also establishing mathematically rigorous principles for prompt construction optimization. Furthermore, we verify the reliability of the explanation, based on which we unify the insights of previous work.


\end{abstract}

\section{Introduction}
\begin{figure*}[t]
    \centering
\includegraphics[width=1.0\linewidth]{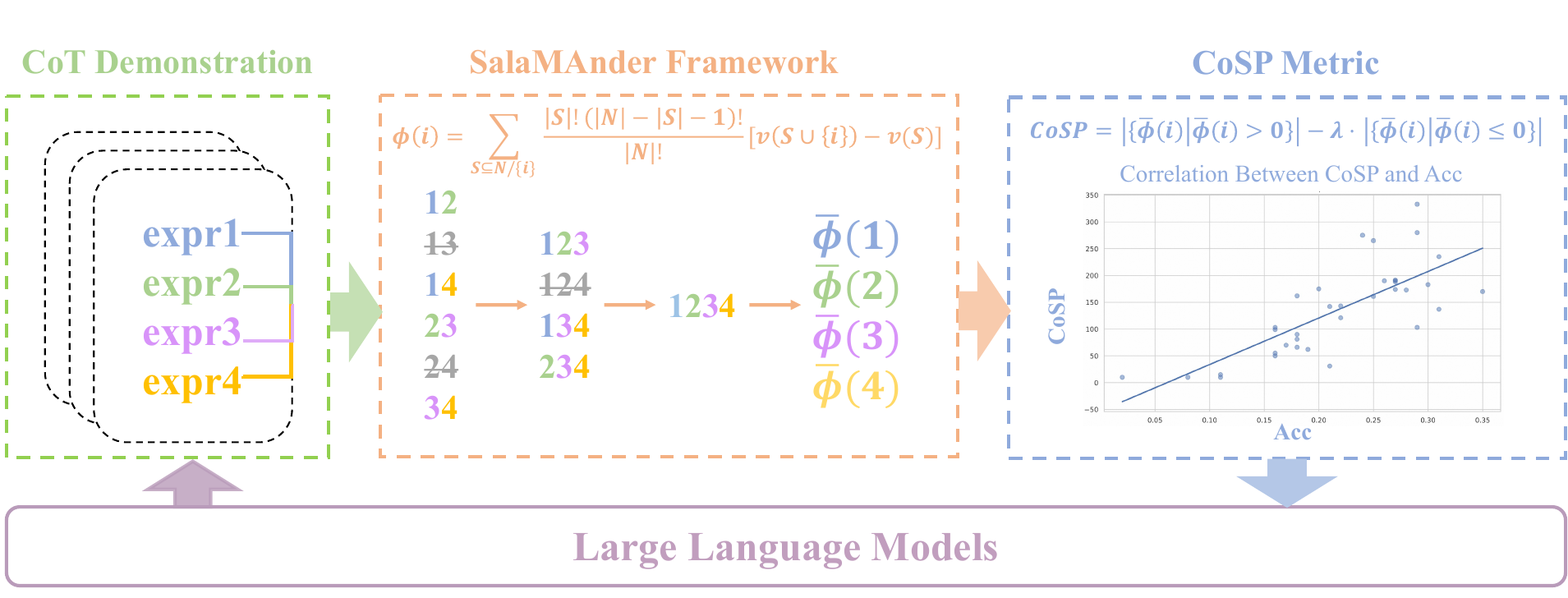}
    \caption{Workflow of the SalaMAnder Framework and CoSP Metric in CoT for LLMs. Initially, the framework proposes an efficient Shapley value algorithm to attribute the contributions of various mathematical expressions. These computed Shapley values are then utilized to derive the CoSP metric. Both theoretical derivations and extensive experiments across multiple models and datasets validate that CoSP exhibits a robust positive correlation with model inference accuracy. This correlation provides a comprehensive explanation of the underlying mechanisms driving CoT behavior in LLMs.}
    \label{fig:frame}
    \vspace{-0.4cm}
\end{figure*}
Chain-of-Thought (CoT) reasoning has elicited powerful mathematical ability within large language models (LLMs) reasoning tasks, ranging from arithmetic problem solving to theorem proving. Despite the substantial improvements, the mechanism of how reasoning steps lead to correct answers remains underexplored, both heuristic speculation\citep{cot_rea1, cot_rea2, cot_rea3, cot_rea4, cot_rea5, cot_rea6} and labor-intensive verification\citep{attn1, attn2, attn3, saliency} lack theoretical investigation.

Prior heuristic-driven approaches analyze the role of different components by defining customized input formats. For instance,~\citet{cot_rea2} and~\citet{cot_rea5} introduce tailored reasoning steps during inference and investigate the impact of step order and length, respectively. While labor-intensive approaches attempt to explain CoT actions through ad hoc trial-and-error adjustments and case-specific manual inspections~\citep{attn1, attn2, attn3, saliency}. There is also a Shapley-value-based method~\citep{tokenshap} analyzing token-level attribution; nevertheless, the exponential computational complexity and indirect value function design hinder it from real-world applications.

In this paper, we propose a unified framework \textbf{SalaMAnder} (short for \textbf{S}h\textbf{a}p\textbf{l}ey-b\textbf{a}sed \textbf{M}athematical Expression \textbf{A}ttribution a\textbf{nd} M\textbf{e}t\textbf{r}ic), introducing two novel ideas for efficient and semantically coherent CoT analysis. First, we denote mathematical expressions as atomic units for Shapley-based attribution, addressing the semantic fragmentation inherent in traditional token-level analyses through component-level decomposition. Then, we develop a novel stratified sampling algorithm, namely \textbf{SalaMA (Shapley-based Mathematical Expression Attribution)} that achieves exponential complexity reduction by decomposing Shapley calculations according to component order, reducing time complexity from $O(2^{n+1})$ to $O(2mn^2)$ while maintaining rigorous theoretical guarantees, where $n$ refers to the number of components and $m$ indicates the number of samples.
To supplement SalaMA, we also develop the \textbf{CoSP} (\textbf{C}ardinality \textbf{o}f \textbf{S}hapley \textbf{P}ositives) metric based on the efficient and semantical Shapley estimation.

The proposed CoSP metric within our SalaMAnder framework formally establishes the monotonic relationship with model performance. Theoretically, we provide a rigorous mathematical analysis of this monotonic relation. Experimentally, we apply SalaMAnder to few-shot learning scenarios, utilizing popular LLMs (LLaMA-2-13B-chat \citep{llama2}, LLaMA-3-8B-Instruct \citep{llama3}, and Qwen2.5-7B-Instruct \citep{qwen25}) tested on various mathematical benchmarks (GSM8K \citep{gsm8k}, MathQA \citep{mathqa}, AQUA \citep{aqua}, MultiArith \citep{multiarith}, and SVAMP \citep{svamp}) to compute the Pearson correlation coefficient. Then we further evaluate the reliability of the explanation results. Last, we present novel insights that not only reinforce the effectiveness of our methods but also integrate and unify previous research.


The contributions of this paper can be summarized as follows:
\begin{enumerate} 
\setlength{\itemsep}{0pt}
\setlength{\topsep}{2pt}
\setlength{\partopsep}{0pt}
    \item[$\bullet$] We propose a unified framework SalaMAnder to establish mathematical expressions as atomic units for Shapley-based attribution, and we develop a novel stratified sampling algorithm SalaMA that achieves exponential complexity reduction while maintaining rigorous theoretical guarantees. 
    \item[$\bullet$] We present the CoSP metric within our SalaMAnder framework, which formally establishes the monotonic relationship with model performance through rigorous covariance analysis, providing mathematical guarantees for the predictive validity.
\end{enumerate}


\section{Related Work}

\paragraph{CoT Methodologies}

CoT prompting, introduced by \citet{cot1}, explicitly guides LLMs to generate intermediate reasoning steps, significantly improving performance on mathematical and symbolic tasks. Subsequent work expanded this paradigm through path optimization (e.g., Least-to-Most prompting decomposes problems into subquestions \citep{cot2}; Progressive-Hint iteratively refines solutions \citep{cot3}), automation (e.g., Automatic CoT generates demonstrations via LLMs \citep{cot4}; Symbolic CoT Distillation transfers CoT ability to smaller models \citep{cot5}), and hybrid approaches (e.g., CoF-CoT combines coarse-to-fine prompting for multi-domain tasks \citep{cot6}; Deductive Verification adds formal consistency checks \citep{cot7}). Despite these advances, most methods rely on heuristic designs without theoretical guarantees, and their efficacy varies significantly across domains—mathematical tasks benefit more from structured CoT than open-ended reasoning.

\paragraph{Mechanistic Studies of CoT Reasoning}
The existing literature on CoT mechanisms unfolds through complementary empirical and theoretical lenses. Empirical studies \citep{cot_rea3, cot_rea4, cot_rea5, cot_rea1, cot_rea6, cot_rea2} have explored various strategies to enhance the robustness, safety, and structural integrity of CoT reasoning. For instance, self-consistency mechanisms \citep{cot_rea3} improve the reliability of reasoning outputs by aggregating multiple reasoning paths, while efforts to mitigate toxicity \citep{cot_rea4} ensure safer commonsense reasoning. Additionally, research on step length \citep{cot_rea5}, step relevance and logical order \citep{cot_rea1}, hidden state dynamics \citep{cot_rea6}, and premise sequence order \citep{cot_rea2} underscores the importance of prompt design and structural factors in optimizing CoT performance. 

Another set of literature attempts to explain CoT through ad hoc trial-and-error adjustments~\citep{attn1, attn2, attn3, saliency}. For instance, 
~\citep{attn2} and ~\citep{saliency} utilize attention maps and saliency score to analyze CoT, respectively. There is also a Shapley-value-based method~\citep{tokenshap} analyzing token-level attribution, nevertheless, the exponential computational complexity and indirect value function design hinder it from real-world applications.



\section{Method}
In this section, we introduce the \textbf{SalaMAnder} framework, designed to explain the mathematical reasoning mechanisms of CoT in LLMs using Shapley values. We introduce our method in three sections: an introduction to Shapley values, the \textbf{SalaMAnder} sparse computation of these values, and the \textbf{CoSP} metric for evaluating CoT reasoning contributions.

\subsection{Preliminary: Shapley Values (Fair Attribution of CoT Constituents)}
Shapley values, originating from cooperative game theory, offer a principled method for fairly distributing the total gains of a coalition among its individual players based on their contributions \citep{shapley}. 

Formally, consider a set of players $ N = \{1, 2, \dots, n\} $ and a reward function $ v: 2^N \rightarrow \mathbb{R} $ that assigns a real-valued payoff to every possible coalition of players. The Shapley value $ \phi_i(v) $ for player $i$ is defined as:
\begin{align*}
    \phi_v(i)=\sum\limits_{S \subseteq N \setminus \{i\}} \dfrac{s! (n - s - 1)!}{n!} \left[ v(S \cup \{i\}) - v(S) \right]
\end{align*}
where $S$ is any subset of $N$ that does not include player $i$, and $s=|S|, n=|N|$ respectively denotes the number of players in subset $S$ and set $N$.

We can further derive from the above expression:
\begin{align}
    \phi(i) &=\dfrac{1}{n}\sum\limits_{S \subseteq N \setminus \{i\}} \dfrac{1}{\binom{n-1}{s}}\left[ v(S \cup \{i\}) - v(S) \right] \notag \\
    &= \dfrac{1}{n}\sum\limits_{r=0}^{n-1} \mathbb{E}_{s=r} \left[ v(S \cup \{i\}) - v(S) \right] \notag \\
    &= \dfrac{1}{n} \phi_{r+1}(i)
    \label{shap_order}
\end{align}
where $\phi_{k}(i) = \mathbb{E}_{s=r} \left[ v(S \cup \{i\}) - v(S) \right]$ denotes the $(r+1)$th order shapley value of component $i$.

Researchers have proven that the Shapley value is a unique unbiased method to fairly allocate overall reward to each player with four properties: linearity, dummy, symmetry, and efficiency \citep{shapley2}. For simplicity, we use $\phi(i)$ by ignoring the superscript of $\phi_v(i)$ in the following manuscript without causing ambiguity.

In our framework, each component of the CoT, such as individual mathematical expressions or a single word, is treated as a player in the cooperative game. The reward function $ v(S) $ corresponds to a performance metric of the LLM (e.g., correctness, or inference logits) when only the components in subset $ S $ are included in the CoT. Consequently, the Shapley value  $\phi(i)$ quantifies the average marginal contribution of each component to the overall reasoning performance across all possible subsets of components.

The feasibility of our method is guaranteed by the non-essential requirement for independence among components in the mathematical definition of the Shapley value and the value function, although low feature independence truly has some problems. But in our specific context, expressions tend to carry distinct semantic roles, with rare high-level redundancy between them. Consequently, the computed Shapley values can effectively reflect the true contribution of each component in the reasoning process.


\subsection{SalaMA: Efficient Sparse Shapley Computation for CoT Components}
Although calculating exact Shapley values for each component presents significant computational challenges, the exponential growth in the number of possible subsets with respect to the number of components renders exact computation infeasible for practical applications. To address the limitation, we propose SalaMA (Shapley-based Mathematical Expression Attribution) mechanism, an efficient algorithm designed to approximate Shapley values with high accuracy while substantially reducing computational overhead.

\paragraph{The Players} We define each player in the game, i.e. each component in the demonstration as a mathematical expression rather than individual words or tokens. This decision is motivated by the observation that single words or tokens can vary in meaning across different contexts, making their attribution inconsistent and less meaningful. Mathematical expressions, in contrast, maintain their semantic integrity across diverse reasoning scenarios, providing a more stable and universally applicable unit for analysis. Additionally, aggregating tokens into coherent mathematical expressions significantly reduces the number of components, thereby mitigating the computational complexity associated with Shapley value calculations. This aggregation not only enhances computational efficiency but also ensures that the attribution analysis remains interpretable and relevant to the model's problem-solving mechanisms.

\paragraph{The Reward Function} We adopt a reward function that combines the model's prediction confidence logits with the correctness of the prediction, formulated as 
\begin{align}\left\{\begin{aligned}
&v(S) = \left( \dfrac{1}{L}\sum_{\ell=1}^L \log p_\theta(y_\ell | S) \right) \cdot \mathbb{I}(y_{\text{pred}}(S) = y^*)\\
&y_{\text{pred}}(S) = \bigoplus_{\ell=1}^L y_\ell(S) 
\end{aligned}\right.\end{align}
where $\frac{1}{L}\sum_{\ell=1}^L \log p_\theta(y_\ell | S) $ represents the average confidence score of the model's prediction by averaging the logits associated with the result tokens generated when including component subset $S$, $\mathbb{I}(\cdot)$ is a binary indicator, and $\bigoplus$ indicates the string concatenation operation.

This formulation ensures that the value function directly reflects the impact of each component on the model's performance, addressing the limitations of alternative metrics such as attention, saliency scores or binary correctness. Attention or saliency scores do not provide a direct attribution to the final outcome and can be complex to interpret \citep{attn1, attn2, attn3, saliency}, while a binary correctness metric lacks the sensitivity needed to capture nuanced contributions. By integrating confidence logits with correctness, the reward function balances sensitivity and direct attribution, facilitating a more accurate and interpretable estimation of each component's contribution.

\paragraph{Efficient Shapley Computation Algorithm}
The proposed algorithm systematically approximates the Shapley values for CoT components through a structured algorithmic workflow. In exact Shapley value computation, for each component $i$, it is necessary to evaluate $v(S\cup \{i\}) - v(S)$ across all subsets $S\subseteq N\{i\}$, leading to a computational complexity of $O(2^{n+1})$, where $n$ is the number of components. This exponential complexity becomes prohibitively expensive as the number of components increases. To mitigate this, SalaMA reduces the number of necessary inferences by employing a stratified sampling approach based on the order of Shapley values.

Specifically, the SalaMA mechanism decomposes the Shapley value calculation by order. For an $r$-th order Shapley value $\phi_r$, SalaMA randomly samples $r-1$ other mathematical expressions from the set $N/\{i\}$. The number of such samples is denoted by $sp$, with a maximum limit of $m$, indicating $sp = \min(m, \binom{n-1}{r-1})$. In the original demonstration, aside from the mathematical expressions, other components (referred to as the "whiteboard") are always present and remain constant across different subsets.

During inference, for each sampled subset $S$ of size 
$r-1$, SalaMA constructs two distinct demonstrations: one containing $S\cup \{i\}$, and another containing $S$, all combined with the whiteboard. These demonstrations are then fed into the model to obtain the corresponding reward functions $v(S\cup \{i\})$ and $v(S)$, respectively. By iterating over multiple orders and different samples within each order, SalaMA aggregates the marginal contributions across various subset configurations. The approximated Shapley value can be derived from Eq.~\eqref{shap_order}:
\begin{align}
    \phi(i) &= \dfrac{1}{n}\sum\limits_{r=0}^{n-1}\mathbb{E}_{s=r}[v(S\cup\{i\} - v(S))]\notag \\
    &= \dfrac{1}{n}\sum\limits_{r=0}^{n-1} \dfrac{1}{m}\sum\limits_{t=1}^m[v(S_t^r\cup \{i\}) - v(S_t^r)]
\end{align}

To further enhance computational efficiency, SalaMA maintains a hash table $\mathcal{H}$ to store and retrieve the results of previously computed subsets $S$. 
This caching mechanism avoids redundant inferences by storing $v(S)$ for each evaluated subset $S$. Consequently, the computational complexity of SalaMA is reduced to $O(2\cdot sp\cdot n^2) \leqslant O(2mn^2)$, which is significantly lower than the exact Shapley value computation's $O(2^{n+1})$. The whole workflow is shown in Algorithm.~\ref{alg:salama}. We also conduct experiments on the computation complexity and error magnitude of Shapley value in Appendix~\ref{sec:compute}, indicating that it is entirely feasible to achieve a trade-off between computational complexity and estimation accuracy with appropriate hyperparameter selection.

\begin{algorithm}[t]
\caption{SalaMA: Sparse Shapley Value Computation}
\label{alg:salama}
\Fn{SalaMA($N, v, n, m$)}{
  Initialize $\phi[i] \gets 0\ (\forall i \in N),\ \mathcal{H} \gets \emptyset$\;
  \ForEach{$i \in N$}{
    \For{$r=1$ \KwTo $n$}{
      $sp \gets \min(m, \binom{n-1}{r-1})$\;
      \For{$s=1$ \KwTo $sp$}{
        $S \gets \text{Sample}(r-1, N\setminus i)$\;
        $v_S \gets \texttt{MemEval}(S, \mathcal{H})$\;  
        $v_{S\cup i} \gets \texttt{MemEval}(S\cup i, \mathcal{H})$\;
        $\phi[i] \mathrel{+}= (v_{S\cup i} - v_S)/(sp \cdot n)$\;
      }
    }
  }
  \Return $\phi$\;
}
\BlankLine
\Proc{\texttt{MemEval}($S, \mathcal{H}$)}{
  \If{$S \notin \mathcal{H}$}{
    $\mathcal{H}[S] \gets v(S)$\;
  }
  \Return $\mathcal{H}[S]$\;
}
\end{algorithm}


\begin{table*}[!t]
    \centering
    \small
    \renewcommand{\arraystretch}{1.2}
    \setlength{\tabcolsep}{5.5pt}
    \begin{tabular}{@{}lcccccccccccc@{}}
        \toprule
        & \multicolumn{4}{c}{LLaMA 2 ($\uparrow$)} & \multicolumn{4}{c}{LLaMA 3 ($\uparrow$)} & \multicolumn{4}{c}{Qwen 2.5 ($\uparrow$)} \\
        \cmidrule(lr){2-5} \cmidrule(lr){6-9} \cmidrule(lr){10-13} 
        & \multicolumn{1}{c}{CoSP-0} & \multicolumn{1}{c}{CoSP-1} & \multicolumn{1}{c}{SSV} & \multicolumn{1}{c}{NoE} & \multicolumn{1}{c}{CoSP-0} & \multicolumn{1}{c}{CoSP-1} & \multicolumn{1}{c}{SSV} & \multicolumn{1}{c}{NoE} & \multicolumn{1}{c}{CoSP-0 } & \multicolumn{1}{c}{CoSP-1} & \multicolumn{1}{c}{SSV} & \multicolumn{1}{c}{NoE} \\
        \midrule
        \rowcolor{mygray} \multicolumn{13}{c}{\textbf{\textit{1-shot}}}\\
        GSM8K  & \textbf{0.76} & 0.65 & 0.32 & 0.76 & 0.70 & 0.18 & -0.14 & \textbf{0.71} & \textbf{0.64} & 0.62 & 0.54 & 0.43\\
        MathQA & 0.44 & \textbf{0.62} & 0.63 & -0.08 & 0.37 & \textbf{0.28} & 0.19 & 0.10 & -0.16 & \textbf{0.28} &  0.11 & -0.22  \\
        AQUA & 0.40 & \textbf{0.46} & 0.44 & -0.31 & -0.21 & \textbf{0.48} & 0.39 & -0.40 & -0.63 & \textbf{-0.03} & -0.03 & -0.67  \\
        MultiArith & \textbf{0.60} & 0.52 & 0.02 & 0.53 & \textbf{0.74} & 0.44 & 0.44 & 0.09 & 0.78 & 0.71 & \textbf{0.80} & -0.04  \\
        SVAMP & \textbf{0.49} & 0.28 & \textbf{0.21} & 0.14 & 0.17 & \bf0.21 & 0.08 & -0.35 & \textbf{0.56} & 0.50 & 0.56 & -0.32  \\
        \rowcolor{mygray} \multicolumn{13}{c}{\textbf{\textit{2-shot}}}\\
        GSM8K  & \textbf{0.75} & 0.35 & 0.14 & 0.75 & \textbf{0.49} & 0.26 & 0.24 & 0.45 & \textbf{0.80} & 0.48 & 0.51 & 0.13\\
        MathQA & 0.36 & \textbf{0.46} & 0.35 & -0.11 & -0.20 & \textbf{0.01} & 0.07 & -0.05 & -0.20 & -0.14 &  \textbf{-0.03} & -0.06  \\
        AQUA & \textbf{0.56} & 0.51 & 0.48 & -0.47 & \textbf{0.09} & -0.04 & -0.22 & -0.50 & 0.22 & 0.52 & \textbf{0.55} & -0.19  \\
        MultiArith & \textbf{-0.04} & -0.07 & -0.20 & -0.31 & \textbf{0.82} & 0.39 & 0.58 & -0.24 & \textbf{0.44} & 0.18 & 0.16 & 0.06  \\
        SVAMP & \textbf{0.23} & 0.05 & -0.13 & -0.02 & \textbf{0.47} & 0.44 & -0.19 & -0.17 & \textbf{0.69} & 0.61 & 0.53 & -0.02  \\
        \rowcolor{mygray} \multicolumn{13}{c}{\textbf{\textit{4-shot}}}\\
        GSM8K  & \textbf{0.77} & 0.61 & 0.12 & 0.52 & 0.26 & \textbf{0.37} & -0.15 & -0.20 & \textbf{0.80} & 0.58 & 0.52 & 0.31\\
        MathQA & \textbf{0.29} & -0.26 & -0.46 & -0.01 & \textbf{0.40} & 0.28 & -0.02 & -0.67 & \textbf{0.18} & -0.33 & -0.52  & 0.14  \\
        AQUA & \textbf{0.80} & 0.77 & -0.10 & -0.11 & -0.08 & \textbf{0.20} & 0.02 & -0.19 & -0.31 & -0.11 & \textbf{-0.05} & -0.43  \\
        MultiArith & \textbf{0.54} & 0.33 & 0.42 & 0.22 & \textbf{0.80} & 0.23 & -0.001 & -0.47 & \textbf{0.67} & 0.51 & 0.24 & -0.44  \\
        SVAMP & \textbf{0.63} & 0.31 & 0.22 & 0.61 & \textbf{0.10} & 0.07 & 0.36 & -0.17 & \textbf{0.22} & -0.03 & -0.14 & -0.13  \\
        \midrule
        \textbf{Average} & \textbf{0.51} & 0.37 & 0.16 & 0.14 & \textbf{0.33} & 0.25 & 0.11 & -0.14 & \textbf{0.31} & 0.29 & 0.25 & -0.10 \\
        \bottomrule
    \end{tabular}
    \caption{The correlation coefficients between different metrics and model inference accuracy across multiple datasets and models of few-shot tasks. For each dataset and each model, the largest correlation is \textbf{bolded}, indicating the best interpretation method. Here we use `LLaMA 2', `LLaMA 3', and `Qwen2.5' in short for LLaMA-2-13B-chat\citep{llama2}, LLaMA-3-8B-Instruct\citep{llama3}, and Qwen2.5-7B-Instruct\citep{qwen25}.}
    \label{tab:cosp}
    \vspace{-0.4cm}
\end{table*}

\subsection{CoSP: Performance-Aligned Causal Explanation Rationale}
\label{sec:cosp}
We introduce CoSP (\textbf{C}ardinality \textbf{o}f \textbf{S}hapley \textbf{P}ositives), a metric defined as the number of expressions within a demonstration that exhibit positive average Shapley values minus a weighted non-positive average Shapley values across multiple experiments.

Formally, for a demonstration comprising a set of $n$ expressions $N$, CoSP is defined as:
\begin{align*}
    CoSP &=|\{\bar{\phi}(i)| \bar{\phi}(i)>0\}| - \lambda \cdot |\{\bar{\phi}(i)| \bar{\phi}(i)\leqslant 0\}|
    \\&=\sum_{i=1}^n \mathbb{I}(\bar{\phi}(i)>0)-\lambda\cdot \mathbb{I}(\bar\phi(i)\leqslant 0)\\
    &=(1+\lambda)\sum_{i=1}^n\mathbb{I}(\bar{\phi}(i)>0)-\lambda n 
\end{align*}
where $\bar{\phi}(i)$ is the average Shapley value of the $i$-th expression, computed over $m$ different problem instances tested using the same demonstration, formulated as $\bar{\phi}(i) = \frac{1}{m}\sum_{k=1}^m\phi^{(k)}(i)$, $\mathbb{I}(\cdot)$ is the indicator function, returning 1 if the condition inside is true and 0 otherwise, and $\lambda >0$ is the penalty severity for the number of expressions with negative Shapley values. And we assume that during the $m$ CoT reasoning precesses, for each expression $i$, there is $\phi^{(k)}(i) \sim \mathcal{N}(\mu_i, \sigma_i^2)$.

A positive average Shapley value ($\bar{\phi}(i)>0$) indicates that the corresponding mathematical expression contributes positively to the model's reasoning performance; conversely, a non-positive one leads to negative contribution or no contribution. Therefore, CoSP comprehensively quantifies the number of expressions that actively enhance or degrade the model's efficacy in solving problems. A higher CoSP suggests that a greater subset of expressions within the CoT is beneficial while a smaller subset harmful, correlating with improved model performance. Specifically, we define CoSP-0 and CoSP-1, with $\lambda$ equals to 0 and 1, respectively.

To substantiate the relationship between CoSP and performance, we formalize the following two theorems under specific statistical assumptions.

\begin{theorem}
Both CoSP-0 and CoSP-1 have positive correlation with the model performance:
\begin{align}
    &\text{Cov}(CoSP, Perf) = (1+\lambda)(\delta_+-\delta_-)\sum_{i=1}^n\text{Var}(X_i)\notag\\
    &\text{Cov}(Perf, CoSP\text{-0}) = (\delta_+-\delta_-)\sum_{i=1}^n\text{Var}(X_i)\notag\\
    &\text{Cov}(Perf, CoSP\text{-1}) = 2(\delta_+-\delta_-)\sum_{i=1}^n\text{Var}(X_i)
    \label{eq:perf}
\end{align}

\label{thm:perf}
\end{theorem}
where the meaning of $\delta_+, \delta_-, X_i$ will be explained in the proof. 

\begin{theorem}
CoSP-0 has a positive correlation with the number of expressions $n$, while CoSP-1 has a negative correlation with $n$:
\begin{align}
        \mathbb{E}[CoSP_{n+1}] &= (1+\lambda)\sum_{i=1}^{n+1}p_i - (n+1)\lambda \notag\\
        &=\mathbb{E}[CoSP_n] + p_{n+1}-\lambda
    \end{align}
\begin{align}
    &\mathbb{E}[CoSP\text{-0}_{n+1}] - \mathbb{E}[CoSP\text{-0}_{n}] = p_{n+1} >0 \notag\\
    &\mathbb{E}[CoSP\text{-1}_{n+1}] - \mathbb{E}[CoSP\text{-1}_{n}] = p_{n+1}-1 <0
\end{align}
    
\label{thm:num}
\end{theorem}

The proof of Theo.~\ref{thm:perf} and Theo.~\ref{thm:num} is applied in Appendix.~\ref{sec:app1}.

The number of expressions $n$ in the CoT is often indicative of the complexity or difficulty of the reasoning task. Generally, increased reasoning difficulty generally leads to better model performance \citep{o1}, provided that the additional complexity is constructively leveraged. Our Theo.~\ref{thm:num} aligns with this observation by showing that a higher number of expressions $n$ results in a higher CoSP-0, which in turn, per Theo.~\ref{thm:perf}, correlates with enhanced model performance. This consistency underscores the validity of CoSP as a metric that not only accounts for the quantity of reasoning steps but also their qualitative impact on model efficacy.

\begin{figure}[!t]
    \centering
    \includegraphics[width=1.0\linewidth]{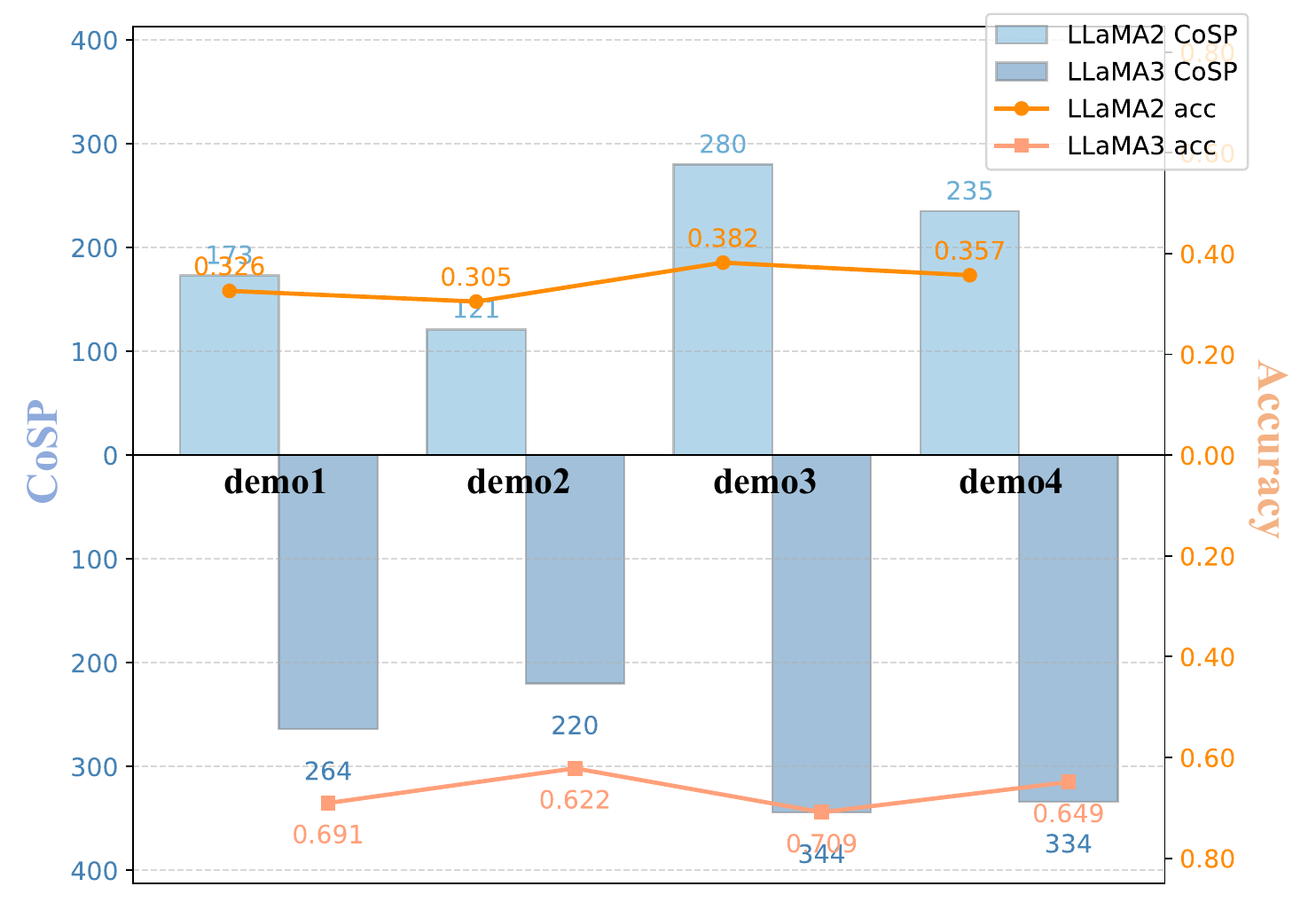}
    \caption{The CoSP-0 value and test accuracy of models. The strong consistency in their variation patterns further confirms the reliability of our explanation results.}
    \label{fig:exp3}
    \vspace{-0.3cm}
\end{figure}

\section{Experiments}
This section presents a comprehensive evaluation of the proposed SalaMAnder framework, demonstrating its applicability across various settings. Appendix~\ref{sec:exp1} describes the experimental settings, and Sec~\ref{sec:exp2} utilizes SalaMAnder in few-shot learning scenarios to assess the validity of our explanation method and metric. In Sec~\ref{sec:exp3}, we further evaluate the reliability of explanation results. Sec~\ref{sec:exp4} presents novel insights that not only reinforce the effectiveness of our methods but also integrate and unify previous research, and Sec~\ref{sec:quality} provides qualitative analysis of the explanation.

Besides, we conduct experiments on the computation complexity and error magnitude of the calculation of Shapley value in Appendix~\ref{sec:compute}, indicating that it is entirely feasible to achieve a trade-off between computational complexity and estimation accuracy, thus guiding the selection of sample num. Appendix~\ref{app:aba} illustrates ablation studies on the hyperparameters and Appendix~\ref{app:enh_aba} presents additional experimental results. And we show the cases used in Sec~\ref{sec:exp4} in Appendix~\ref{sec:app2}, more cases in Appendix~\ref{sec:cases}, and the qualitative analysis cases in Appendix~\ref{sec:quality_cases}.
\begin{table*}[t]
    \centering
    \small
    \renewcommand{\arraystretch}{1.2}
    \setlength{\tabcolsep}{6pt}
    \begin{tabular}{lcccccccccccc}
        \toprule
        & \multicolumn{4}{c}{1-shot ($\uparrow$)} & \multicolumn{4}{c}{2-shot ($\uparrow$)} & \multicolumn{4}{c}{4-shot ($\uparrow$)} \\
        \cmidrule(lr){2-5} \cmidrule(lr){6-9} \cmidrule(lr){10-13} 
        & \multicolumn{1}{c}{CoSP-0} & \multicolumn{1}{c}{CoSP-1} & \multicolumn{1}{c}{SSV} & \multicolumn{1}{c}{NoE} & \multicolumn{1}{c}{CoSP-0} & \multicolumn{1}{c}{CoSP-1} & \multicolumn{1}{c}{SSV} & \multicolumn{1}{c}{NoE} & \multicolumn{1}{c}{CoSP-0} & \multicolumn{1}{c}{CoSP-1} & \multicolumn{1}{c}{SSV} & \multicolumn{1}{c}{NoE} \\
        \midrule
        GSM8K  & \textbf{0.70} & 0.48 & 0.24 & 0.63 & \textbf{0.68} & 0.36 & 0.33 & 0.44 & \textbf{0.61} & 0.52 & 0.16  & 0.21 \\
        MathQA & 0.22 & \textbf{0.39} & 0.31 & -0.07 & -0.01 & 0.11 & \textbf{0.13} & -0.07 & \textbf{0.29} & -0.10 & -0.33 & -0.18  \\
        AQUA & -0.15 & \textbf{0.30} & 0.27 & -0.46 & 0.29 & \textbf{0.33} & 0.27 & -0.39 & 0.14 & \textbf{0.29}  & -0.04 & -0.24  \\
        MultiArith & \textbf{0.71} & 0.56 & 0.02 & 0.42 & \textbf{0.41} & 0.17 & 0.18 & -0.16 & \textbf{0.64} & 0.36  & 0.22  & -0.23  \\
        SVAMP & \textbf{0.41} & 0.33 & 0.28 & -0.18 & \textbf{0.46} & 0.37 & 0.07 & -0.07 & \textbf{0.32} & 0.12 & 0.15  & 0.10  \\
        \bottomrule
    \end{tabular}
    \vspace{-0.2cm}
    \caption{The correlation coefficients averaged among various models in few-shot tasks. For each dataset, the largest correlation is \textbf{bolded}, indicating the best interpretation method.}
    \label{tab:cosp_avg}
    \vspace{-0.4cm}
\end{table*}



\subsection{Attribution Validity: CoSP Metric Verification in Few-Shot Learning}
\label{sec:exp2}
To evaluate the practical applicability of the proposed SalaMA method and the CoSP metric, we applied them to few-shot learning scenarios across multiple mathematical datasets and foundational language models to assess the correlation between CoSP and model performance (accuracy), thereby validating the effectiveness of our framework.

We meticulously constructed demonstrations to ensure a uniform distribution of mathematical expressions. Specifically, for one-shot learning tasks, we constructed demonstrations by selecting 35 question-answer (Q-A) pairs from the training sets of the GSM8K, MathQA, and AQUA datasets. Because the MultiArith and SVAMP datasets include answers composed solely of single mathematical expressions, we instead selected 35 Q-A pairs from the GSM8K dataset to serve as demonstrations. These one-shot demonstrations were evenly distributed, with five Q-A pairs each containing between one and seven mathematical expressions. For 2-shot demonstrations, the total number of expressions ranged from 2 to 10, resulting in 14 unique demonstrations by accounting for multiple combinations where applicable (e.g., a total of 6 expressions could be achieved by combinations 2+4 or 3+3). 4-shot demonstrations contained 4-16 total expressions, with one unique combination retained per expression count to minimize computation, producing 13 distinct demonstration sets. This methodology ensured that both one-shot and few-shot demonstrations maintained a balanced and uniform distribution of mathematical expressions, thereby isolating the effect of expression quantity on model performance.

We then utilize the proposed SalaMA method to few-shot learning to get various metrics: CoSP-0, CoSP-1, SSV (the sum of averaged shapley value, i.e. $\sum_{i=1}^n\bar\phi(i)$), NoE(number of expressions, i.e. $n$). The correlations of these metrics and model inference accuracy across diverse datasets and models in 1, 2, 4-shot scenarios are shown in Tab.~\ref{tab:cosp}, and Tab.~\ref{tab:cosp_avg} record the correlations averaged among different models.

\begin{figure*}[!t]
       \centering
       \begin{subfigure}[b]{0.235\textwidth}
           \centering
           \includegraphics[width=\textwidth]{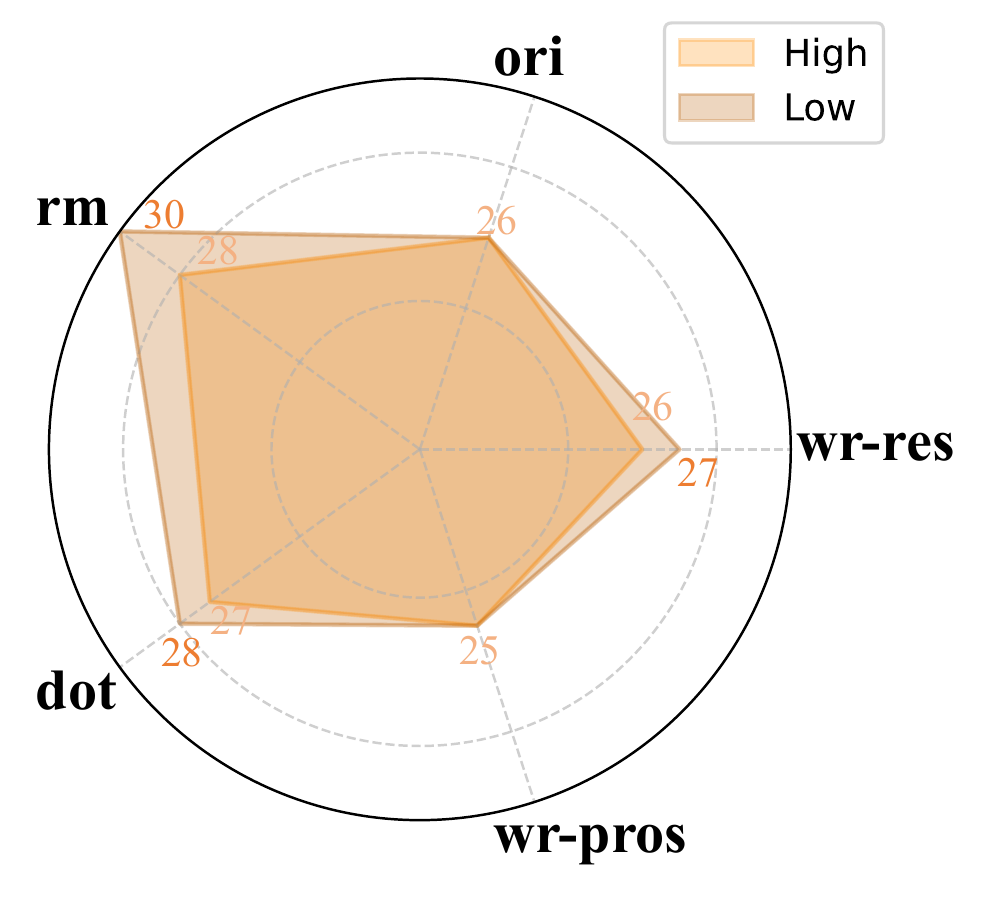}
           \caption{LLaMA2 demo1}
       \end{subfigure}%
       \hfill
       \begin{subfigure}[b]{0.235\textwidth}
           \centering
           \includegraphics[width=\textwidth]{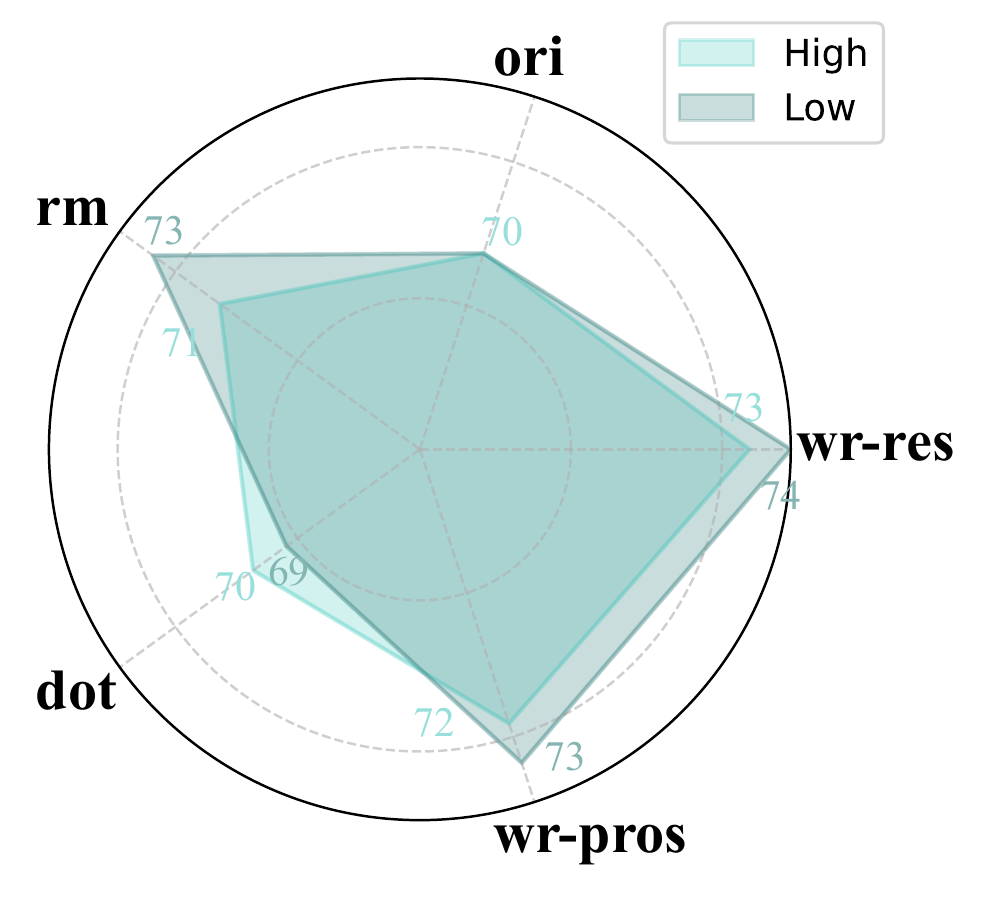}
           \caption{LLaMA3 demo1}
       \end{subfigure}%
       \hfill
       \begin{subfigure}[b]{0.235\textwidth}
           \centering
           \includegraphics[width=\textwidth]{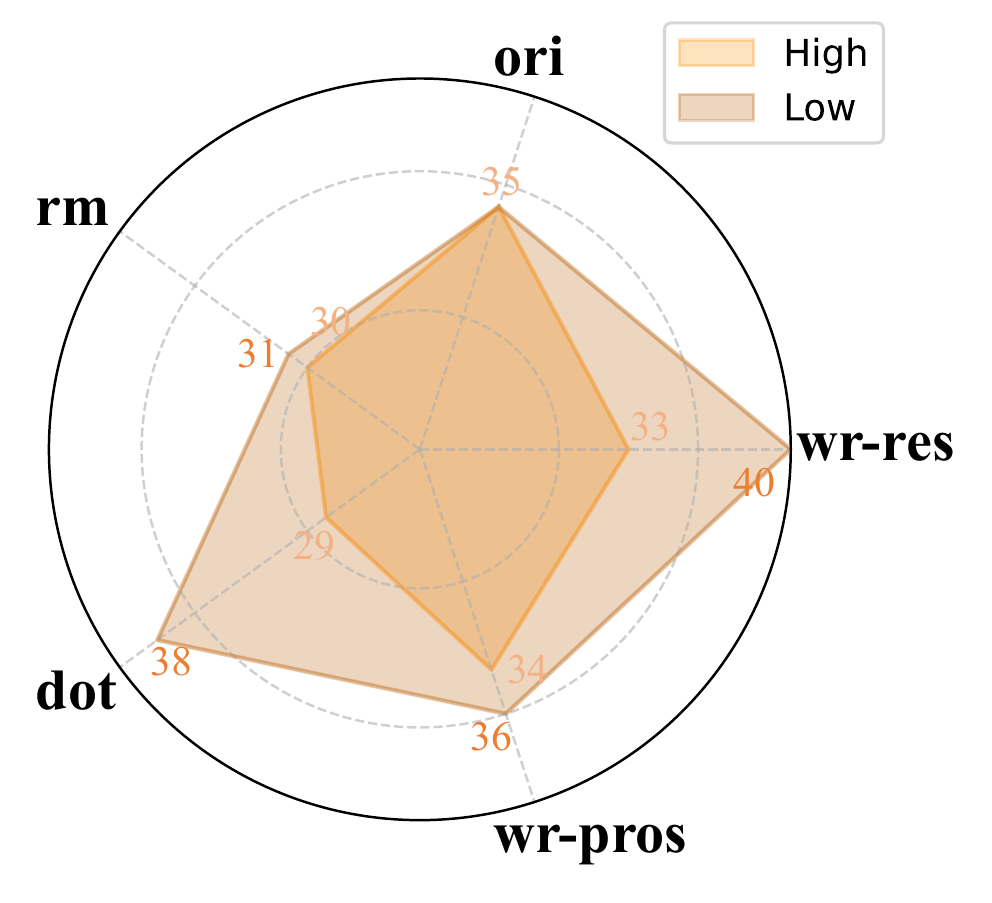}
           \caption{LLaMA2 demo2}
       \end{subfigure}%
       \hfill
       \begin{subfigure}[b]{0.235\textwidth}
           \centering
           \includegraphics[width=\textwidth]{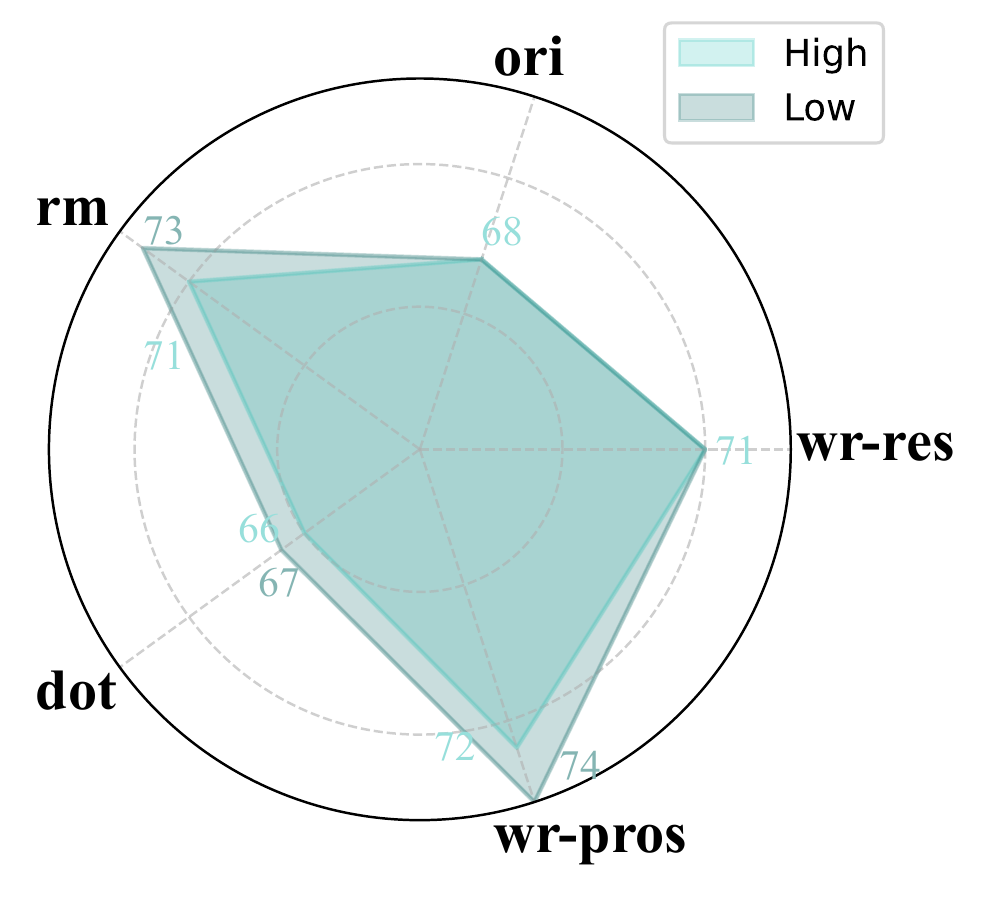}
           \caption{LLaMA3 demo2}
       \end{subfigure}%
        \vspace{-0.3cm}
       \caption{Accuracy of demonstrations for low and high CoSP-0 expressions after four types of modifications in the test set across different models and demos: (a) LLaMA2-demo1, (b) LLaMA3-demo1, (c) LLaMA2-demo2, and (d) LLaMA3-demo2. The observed results indicate that the accuracy curve for low CoSP-0 expressions encompasses that for high CoSP-0 expressions in almost all scenarios, highlighting that alterations on low CoSP-0 expressions yield overall better performance outcomes compared to alterations on high CoSP-0 expressions.}
       \label{fig:insight}
       \vspace{-0.5cm}
   \end{figure*}
Observed from Tab.~\ref{tab:cosp}, CoSP-0 is the best interpretation metric for all models, and the interpretation validity of CoSP-0/CoSP-1 is much better than the other metrics. According to Tab.~\ref{tab:cosp_avg}, CoSP-0 serves as the best interpretation metric for GSM8K, MultiArith, and SVAMP, while CoSP-1 for AQUA. For MathQA, CoSP-0 serves as the best interpretation metric in 1 or 2-shot learning, while CoSP-1 the best in 4-shot learning.

\subsection{Explanation Reliability: Large-Scale Testing Assessment of CoSP Explanations}
\label{sec:exp3}
To further assess the reliability of our CoSP explanations, we conducted comprehensive validation experiments using the entire test set of the GSM8K dataset with both the LLaMA 2 and LLaMA 3 models. This focused approach ensures generality while maintaining computational feasibility. We selected four demonstrations for each model where the CoSP-0 scores for LLaMA 2 is 173, 121, 280, 235, while for LLaMA 3 is 264, 220, 344, 334.

The experimental outcomes consistently demonstrated a strong positive correlation between CoSP-0 scores and model accuracy for both LLaMA 2 and LLaMA 3 according to Fig.~\ref{fig:exp3}, with the experimental results of MathQA and AQUA shown in Fig.~\ref{fig:enh_mathqa_aqua} in Appendix~\ref{app:enh_aba}. Specifically, for LLaMA 2, the demonstration with a CoSP-0 score of 280 achieved the highest accuracy, followed by demonstrations with scores of 235, 173, and 121, in descending order of performance. Similarly, for LLaMA 3, the demonstration with a CoSP-0 score of 344 yielded the highest accuracy, followed by those with scores of 334, 264, and 220. This consistent pattern across both models indicates that demonstrations with higher CoSP-0 scores significantly enhance the reasoning capabilities of the models, while those with lower scores contribute less effectively.

To be mentioned, the strong consistency in CoSP-0 and model accuracy not only confirms the reliability of the explanation results provided by SalaMAnder, but also reveals a potential application in the systematic selection of few-shot demonstrations, rather than random sampling.

\vspace{-0.2cm}
\subsection{Analytical Extensibility: Discovery of Novel Insights in CoT}
\label{sec:exp4}
Building upon our previous findings that high CoSP expressions contribute maximally, while low ones contribute minimally to model reasoning, we sought to uncover novel insights into the dynamics of CoT reasoning processes. Specifically, we applied four distinct altering to the expression with the highest and lowest CoSP-0 to assess their impact on model performance. 1) Removed the expression. 2) Replaced the expressions with non-informative placeholders, i.e., `...'. 3) Introduced calculation errors, for example, converting from `2 + 3 = 5' to `2 + 3 = 6'. 4) Introduced process errors, for example, converting from `2 + 3 = 5' to `4 + 7 = 11'. And we selected two demonstrations and conducted these experiments on GSM8K datasets, with both the LLaMA 2 and LLaMA 3 models. The original demonstration is presented in Appendix~\ref{sec:app2}, where different expressions of CoSP in different colors. Additional experiments on MathQA and AQUA are illustrated in Appendix~\ref{app:enh_aba} and more cases are shown in Appendix~\ref{sec:cases} for reference.

Figures \ref{fig:insight} depict the effect of these alterations on the accuracy of the test set for low and high CoSP expressions across different demonstrations and models. It was consistently observed across almost all experiments that the performance curves for low CoSP expressions encapsulated those for high CoSP expressions.

The results suggest that modifications to low CoSP expressions lead to better performance outcomes compared to modifications to high CoSP expressions. This finding further corroborates our initial hypothesis: low CoSP expressions exert minimal influence on model reasoning, whereas high ones significantly contribute. 


Additionally, our experimental findings reveal several intriguing phenomena. Notably, the removal of certain expressions, the substitution of expressions with non-informative filler tokens (such as `...'), and the introduction of errors in either the result or process of expressions do not necessarily lead to significant degradation in model performance. This outcome resonates with prior studies\citep{cot_rea6, cot_rea1}.

\subsection{Qualitative Analysis}
\label{sec:quality}
We qualitatively illustrate why some demonstrations have higher CoSP values and are beneficial for model reasoning, while others are not. We analyze both the entire demonstration level and the individual expression level.

As for the CoSP of the entire demonstration, consider Example 1 and Example 2: the first contains more expressions and exhibits a richer logical structure, thus providing more informative signals for the model’s reasoning. The second involves simpler computations (only division and comparison operations) and fails to convey a meaningful reasoning pattern, resulting in a smaller positive contribution. These examples can be found in Appendix~\ref{sec:quality_cases}.

For the individual expression level, consider Example 3, where the expression $6/3=2$ has a low CoSP value, whereas $=>(x_3 + x_4 + x_5 + x_6)/4=85$ has a high one. The former computes an irrelevant intermediate variable (in this case, one third of the strings), which does not contribute meaningfully to the final answer. In contrast, the latter is directly linked to the final solution and builds upon previous computations, making it highly relevant to the reasoning process, thus yielding a higher CoSP. Example 3 is shown below:

\begin{tcolorbox}[
    enhanced,
    breakable,
    width=\columnwidth,
    arc=3mm,
    boxrule=1pt,
    left=2mm,
    right=2mm,
    top=1mm,
    bottom=1mm,
    fontupper=\ttfamily,
    before skip=5mm,
    after skip=5mm,
    title=Example3, 
    coltitle=white, 
    colbacktitle=black, 
    fonttitle=\bfseries 
    ]
    Question: \\
    the average length of 6 strings is 80 cm. if the average length of one third of the strings is 70 cm, what is the average of the other strings ? a ) 75. , b ) 85. , c ) 90. , d ) 94. , e ) 100. \\
    Answer: \\
    edit : given \colorbox{blockgreen}{( x1 + x2 ... + x6 ) / 6 = 80} \colorbox{blockgreen}{( x1 + x2 ... + x6 ) = 480} - - > eq 1 . now given avg length of one third strings is 70 . that means out \colorbox{blockblue}{6 / 3 = 2} strings. let the avg length of two strings be \colorbox{blockgreen}{( x1 + x2 ) / 2 = 70} . \colorbox{blockgreen}{( x1 + x2 ) = 140}. - - > eq 2 . now we are asked to find the average of the remaining i . e . \colorbox{blockgreen}{( x3 + x4 + x5 + x6 )} substitute eq 2 in eq 1 then we get\colorbox{blockgreen}{ 140 + x3 + x4 + x5 + x6 = 480} = > \colorbox{blockgreen}{x3 + x4 + x5 + x6 = 340} now divide 340 by 4 we get 85 . \colorbox{blockorange}{=> ( x3 + x4 + x5 + x6 ) / 4 = 85} = avg length of remaining strings . the correct option is b. 
\end{tcolorbox}

\vspace{-0.2cm}
\section{Conclusion}
\vspace{-0.2cm}
In this paper, we propose \textbf{SalaMAnder}, a novel framework for understanding and optimizing Chain-of-Thought (CoT) reasoning in large language models (LLMs). By introducing a theoretically grounded methodology based on Shapley value attribution and developing the \textbf{CoSP (Cardinality of Shapley Positives)} metric, we have established a mathematically rigorous approach to quantifying component-level contributions in CoT reasoning. 
 Extensive validation across various LLM models and mathematical benchmarks demonstrates that the CoSP metric within our SalaMAnder framework strongly and monotonically correlates with model performance. This correlation not only theoretically explains the empirical success of the existing few-shot CoT but also provides rigorous guidelines for optimizing prompt construction. Furthermore, it can be utilized to discover novel insights resonating with prior studies.

\section*{Limitations}
While SalaMAnder is theoretically a general approach, we are currently focusing on mathematical reasoning problems because they are highly representative of few-shot CoT reasoning. In the future we aim to expand the application of SalaMAnder to a broader array of tasks. 

Due to computational resource constraints, our experiments are currently confined to LLMs with a parameter scale between 7 billion and 13 billion. 

\section*{Acknowledgments}
This work was supported by Alibaba Research Intern Program.

\bibliography{custom}

\clearpage
\appendix

\section{The proof of Theorems}
\label{sec:app1}
We have three assumptions necessary for the proof:
\begin{enumerate}
    \item The positive contribution of any expression has a significant lower bound:
    \begin{align*}
        \exists ~ &\delta_+>0, ~s.t.\\
        &\mu_i > \delta_+\cdot \mathbb{I}(\mu_i>0)
    \end{align*}
    \item The non-positive contribution of any expression has a lower bound:
    \begin{align*}
        \exists ~ &\delta_-<0, ~s.t.\\
        &\mu_i>\delta_-\mathbb{I}(\mu_i\leqslant 0) = \delta_-\cdot (1-\mathbb{I}(\mu_i>0))
    \end{align*}
    \item The contributions of different expressions are mutually independent when applied to different problems:
    \begin{align*}
        &\text{Cov}(\phi^{(k)}(i), \phi^{(l)}(j)) = 0 \\
        &(\forall~i\neq j, 1\leqslant k, l\leqslant m, k\neq l)
    \end{align*}
\end{enumerate}

Here is the proof of Theo.~\ref{thm:perf}:
\begin{proof}
As illustrated in Sec.~\ref{sec:cosp}:
\begin{align*}
    &\phi^{(k)}(i)\sim \mathcal{N}(\mu_i, \sigma_i^2)\\
    &\bar\phi(i) = \dfrac{1}{m}\sum_{k=1}^m \phi^{(k)}(i)\stackrel{m\to \infty}{\longrightarrow} \mu_i
\end{align*}
To simplify the expression, we define a positive contribution indicator $X_i = \mathbb{I}(\bar\phi(i)>0)$. Thus:
\begin{align}
    CoSP &= \sum_{i=1}^nX_i - \lambda \sum_{i=1}^n(1-X_i)\notag\\
    &= (1+\lambda)\sum_{i=1}^nX_i - n\lambda
\end{align}

And we define the model performance $Perf$ by summing the expected shapley value of all expressions:
\begin{align}
    Perf = \sum_{i=1}^n \mathbb{E}[\phi(i)] = \sum_{i=1}^n \mu_i
\end{align}
Thus we can further derive the expression of $Perf$:
\begin{align}
    Perf &= \sum_{i\in S_+}\mu_i + \sum_{i\notin S_+}\mu_i > \sum_{i\in S_+}\delta_+ + \sum_{i\notin S_+}\delta_- \notag \\
    &= \sum_{i=1}^n \delta_+\mathbb{I}(\mu_i>0) + \sum_{i=1}^n\delta_-\mathbb{I}(\mu_i\leqslant 0)\notag \\
    &=\sum_{i=1}^n \delta_+\mathbb{I}(\mu_i>0) + \sum_{i=1}^n\delta_-(1-\mathbb{I}(\mu_i>0))\notag \\
    &=n\delta_- + \sum_{i=1}^n(\delta_+-\delta_-)\cdot \mathbb{I}(\mu_i>0)\notag \\
    &=n\delta_- + (\delta_+-\delta_-)\cdot \dfrac{CoSP+n\lambda}{1+\lambda}
\end{align}
indicating a linear functional relationship between a lower bound of model performance and $CoSP$.

And the coveriance between $\mu_i$ and $X_i$ is:
\begin{align*}
    \text{Cov}(\mu_i, X_i) = \mathbb{E}[\mu_iX_i] - \mathbb{E}[\mu_i]\mathbb{E}[X_i]
\end{align*}
where $\mu_i > \delta_+X_i + \delta_-(1-X_i)$ based on the first two assumptions.

We define a residual item $\epsilon_i>0$, $s.t.:$
\begin{align*}
    \mu_i = \delta_+X_i + \delta_-(1-X_i) + \epsilon_i
\end{align*}

Then
\begin{align*}
    \mathbb{E}[\mu_iX_i] &= \delta_+\mathbb{E}[X_i^2] + \delta_-\mathbb{E}[(1-X_i)X_i] + \mathbb{E}[\epsilon_i X_i]\\
    &=\delta_+ + \mathbb{E}[\epsilon_i X_i]
\end{align*}
The second equation is because $X_i(1-X_i)=0$.

And
\begin{align*}
    \mathbb{E}[\mu_i] = \delta_+\mathbb{E}[X_i] + \delta_-\mathbb{E}[1-X_i] + \mathbb{E}[\epsilon_i]
\end{align*}

Thus
\begin{align*}
    \text{Cov}(\mu_i, X_i) = &\delta_+\mathbb{E}[X_i^2] + \mathbb{E}[\epsilon_iX_i] - \delta_+ \mathbb{E}^2[X_i] - \\
    &\delta_-\mathbb{E}[X_i]\mathbb{E}[1-X_i] + \mathbb{E}[X_i]\mathbb{E}[\epsilon_i]
\end{align*}
Since $\mathbb{E}[X_i] = \mathbb{E}[X_i^2]$, and $\mathbb{E}[1-X_i] = 1-\mathbb{E}[X_i]$, then
\begin{align*}
    \mathbb{E}[X_i]\mathbb{E}[1-X_i] &= \mathbb{E}[X_i](1-\mathbb{E}[X_i])\\
    &=\mathbb{E}[X_i] - \mathbb{E}^2[X_i]\\
    &=\mathbb{E}[X_i^2] - \mathbb{E}^2[X_i]\\
    &=\text{Var}(X_i)
\end{align*}

Then
\begin{align}
    \text{Cov}(\mu_i, X_i) = (\delta_+-\delta_-)\text{Var}(X_i) + \text{Cov}(\epsilon_i, X_i)
\end{align}

Based on the third assumption, we have:

\begin{align*}
    &\text{Cov}(Perf, CoSP) = \sum_{i=1}^n\sum_{j=1}^n \text{Cov}(\mu_i, (1+\lambda)X_j-\lambda)\\
    &= \sum_{i=1}^n\text{Cov}(\mu_i, (1+\lambda)X_i-\lambda) \\
    &= (1+\lambda)\sum_{i=1}^n\text{Cov}(\mu_i, X_i)\\
    &= (1+\lambda)\left[(\delta_+-\delta_-)\sum_{i=1}^n\text{Var}(X_i) + \sum_{i=1}^n\text{Cov}(\epsilon_i, X_i)\right]
\end{align*}

And since the residual $\epsilon_i$ has little relevance with $X_i$, the sum of the covariance tends to $0$. Thus
\begin{align}
    \text{Cov}(Perf, CoSP) &= (1+\lambda)(\delta_+-\delta_-)\sum_{i=1}^n\text{Var}(X_i) \notag\\
    &>0
\end{align}

Specifically, we define CoSP-0 and CoSP-1, with $\lambda$ equals to 0 and 1, respectively. Then
\begin{align}
    \text{Cov}(Perf, CoSP\text{-0}) &= (\delta_+-\delta_-)\sum_{i=1}^n\text{Var}(X_i)\\
    \text{Cov}(Perf, CoSP\text{-1}) &= 2(\delta_+-\delta_-)\sum_{i=1}^n\text{Var}(X_i)
\end{align}

Thus $CoSP$ has a positive correlation with model performance. 

\hfill $\square$

\end{proof}

Here is the proof of Theo.~\ref{thm:num}:
\begin{proof}
    Since $X_i = \mathbb{I}(\bar\phi(i)>0)$, then $X_i$ follows a Bernoulli distribution:
    \begin{align}
        p_i = P(X_i=1) = \Phi(\dfrac{\mu_i}{\sigma_i})
    \end{align}
    where $\Phi(\cdot)$ is the standard normal distribution cumulative function.

    Thus the expected value of $CoSP$ with $n$ expressions is:
    \begin{align}
        \mathbb{E}[CoSP_n] &= (1+\lambda)\sum_{i=1}^n\Phi(\dfrac{\mu_i}{\sigma_i})-n\lambda\notag\\
        &=(1+\lambda)\sum_{i=1}^np_i-n\lambda
    \end{align}


    Thus the expected value of $CoSP$ with $n+1$ expressions is:
    \begin{align}
        \mathbb{E}[CoSP_{n+1}] &= (1+\lambda)\sum_{i=1}^{n+1}p_i - (n+1)\lambda \notag\\
        &=\mathbb{E}[CoSP_n] + p_{n+1}-\lambda
    \end{align}

    Therefore, CoSP-0 increases monotonically with the number of expressions $n$, while CoSP-1 decreases monotonically with $n$.

    \hfill $\square$
\end{proof}

\newpage
\section{Experimental Settings}
\label{sec:exp1}
To evaluate the effectiveness of the proposed SalaMA method and the CoSP metric, we conducted experiments using three foundational large language models and five representative mathematical datasets. The selected models, LLaMA-2-13B-chat \citep{llama2}, LLaMA-3-8B-Instruct \citep{llama3}, and Qwen2.5-7B-Instruct \citep{qwen25} were drawn from various model families, each featuring distinct architectures and parameter sizes. This ensures that our analysis of CoSP and SalaMA is broadly applicable across different model paradigms.

For the datasets, we utilized GSM8K \citep{gsm8k}, MathQA \citep{mathqa}, AQUA \citep{aqua}, MultiArith \citep{multiarith}, and SVAMP \citep{svamp}. These datasets were selected for their representativeness in the mathematical question-answering domain, encompassing a range of difficulties where MathQA and AQUA are approximately equivalent and more challenging than GSM8K, which is in turn more difficult than MultiArith and SVAMP. Specifically, GSM8K consists of grade-school level math problems, MathQA includes complex multi-step reasoning questions, AQUA focuses on arithmetic and algebraic tasks, MultiArith provides multi-step arithmetic word problems, and SVAMP introduces adversarial variations to traditional arithmetic problems. This selection ensures comprehensive coverage of various aspects and complexities inherent in mathematical QA tasks.

\newpage
\section{The Trade-off Between Computation Complexity and Error Magnitude}
\label{sec:compute}
As illustrated before, The computational complexity of the Shapley value is $O(2^{n+1})$, while the complexity of our proposed SalaMa method is $O(2mn^2)$ where $m$ denotes the number of samples, and $n$ indicates the number of mathematical expressions. We evaluate the model inference cost and the relative error between the estimated and true Shapley values under different sampling settings. We randomly select a demonstration with $n=8$ to illustrate the trade-off between efficiency and accuracy, where the maximum number of combinations at each order is $\binom{7}{3}=35$ according to Eq.~\eqref{shap_order}. 
\begin{table}[!h]
    \centering
    \small
    \renewcommand{\arraystretch}{1.2}
    \setlength{\tabcolsep}{5.5pt}
    \begin{tabular}{lcccc}
        \toprule
        $m$ & 5 & 15 & 25 & 35 \\
        \midrule
        error(\%) & 62 & 45 & 12 & 0\\
        \bottomrule
    \end{tabular}
    \caption{The computation complexity and relative error of Shapley value.}
    \label{tab:compute}
\end{table}

As shown in Tab.~\ref{tab:compute}, it is entirely feasible to achieve a trade-off between computational complexity and estimation accuracy by selecting appropriate hyperparameters. For example, setting the sample number to 25 allows us to significantly reduce the computational cost while maintaining high precision in Shapley value estimation.

\newpage
\section{Ablation Study on Hyperparameters}
\label{app:aba}
In this section, we conduct ablation studies on the hyperparameter $\lambda$, which indicates the penalty to the mathematical expressions of negative contribution. $\lambda=0$ shows no penalty and only encourages positive contributions, while $\lambda=1$ demonstrates that equal attention is given to both positive and negative contributions. Thus a feasible value of $\lambda$ is in $[0,1]$. Here we conduct this ablation study on LLaMA2-13B on various mathematical datasets, with the same experimental setup as in Sec~\ref{sec:exp2}.

\begin{table}[!h]
    \centering
    \small
    \renewcommand{\arraystretch}{1.2}
    \setlength{\tabcolsep}{5.5pt}
    \begin{tabular}{@{}lcccc@{}}
        \toprule
        & \multicolumn{4}{c}{LLaMA 2 ($\uparrow$)} \\
        \cmidrule(lr){2-5} 
        & \multicolumn{1}{c}{CoSP-0} & \multicolumn{1}{c}{CoSP-0.5} & \multicolumn{1}{c}{CoSP-0.8} & \multicolumn{1}{c}{CoSP-1} \\
        \midrule
        GSM8K  & 0.76 & \bf0.77 & 0.75 & 0.65 \\
        MathQA & 0.44 & 0.58 & 0.59 & \bf0.62 \\
        AQUA & 0.40 & 0.49 & \bf0.53 & 0.46 \\
        MultiArith & \bf0.60 & 0.60 & 0.54 & 0.52\\
        SVAMP & \textbf{0.49} & 0.34 & 0.27 & 0.28\\
        \bottomrule
    \end{tabular}
    \caption{The correlation coefficients between different metrics and model inference accuracy across multiple datasets on LLaMA2-13B-chat on 1-shot task. }
    \label{tab:cosp}
    \vspace{-0.4cm}
\end{table}

As can be observed, when $\lambda$ is set within the range $(0, 1)$, the correlation between CoSP-$\lambda$ and model accuracy lies between that of CoSP-0 and CoSP-1 on almost all datasets, and it changes almost monotonically with $\lambda$. This suggests a smooth transition in the positive and negative contributions as the penalty weight is adjusted, further supporting the robustness of our approach under different weighting schemes.

\newpage
\section{Additional Experiments}
\label{app:enh_aba}
Additional experiments of Sec~\ref{sec:exp3} and Sec~\ref{sec:exp4} are shown here.

\begin{figure}[!h]
       \centering
       \begin{subfigure}[b]{1.0\linewidth}
           \centering
           \includegraphics[width=\linewidth]{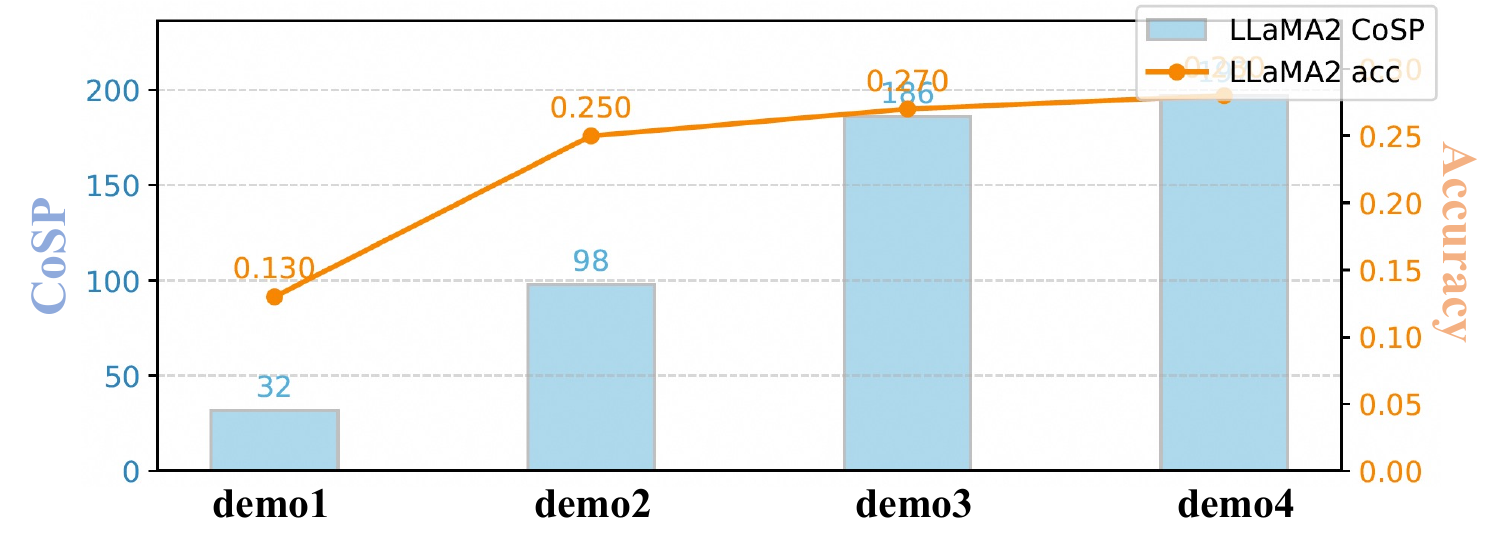}
           \caption{LLaMA2 on MathQA}
       \end{subfigure}%
       \\
       \begin{subfigure}[b]{1.0\linewidth}
           \centering
           \includegraphics[width=\textwidth]{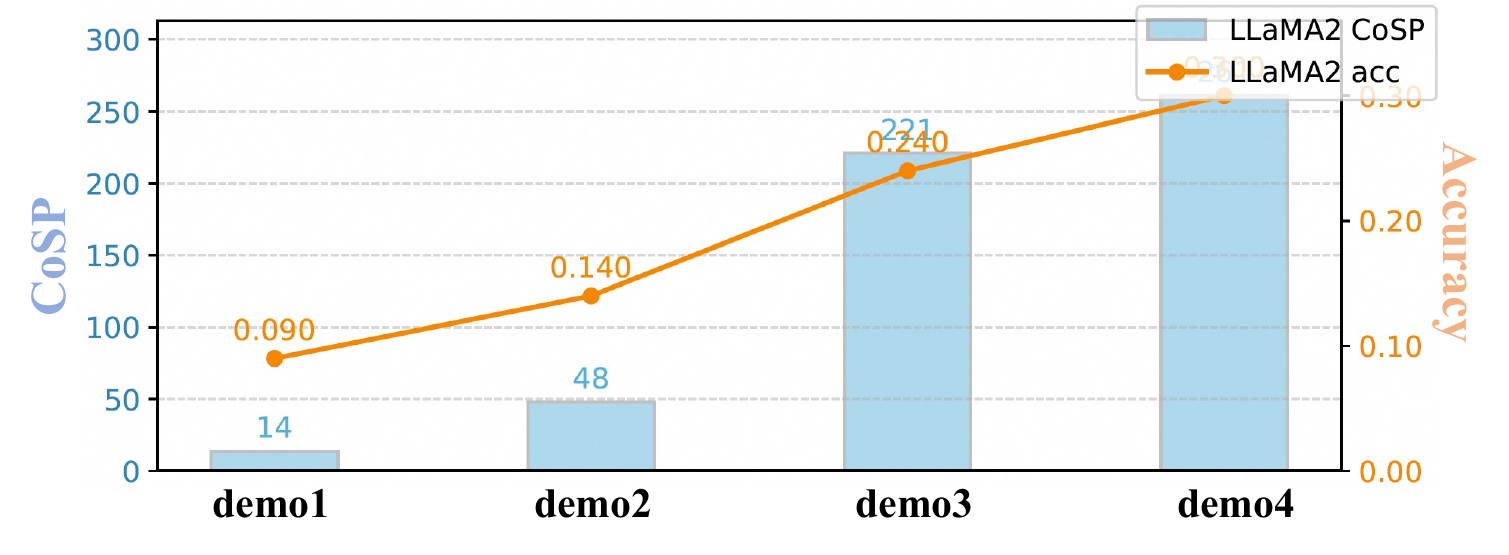}
           \caption{LLaMA2 on AQUA}
       \end{subfigure}%
       \caption{The CoSP-0 value and test accuracy of models. (a) LLaMA2 on MathQA, (b) LLaMA on AQUA.}
       \label{fig:enh_mathqa_aqua}
   \end{figure}

Fig~\ref{fig:enh_mathqa_aqua} exhibits the results of LLaMA2 conducted on MathQA and AQUA, with the same setups as Sec~\ref{sec:exp3}.
Fig.~\ref{fig:enh_mathqa_aqua} illustrates strong consistency in the model performance and CoSP-0.

\begin{figure}[!h]
       \centering
       \begin{subfigure}[b]{0.24\textwidth}
           \centering
           \includegraphics[width=\textwidth]{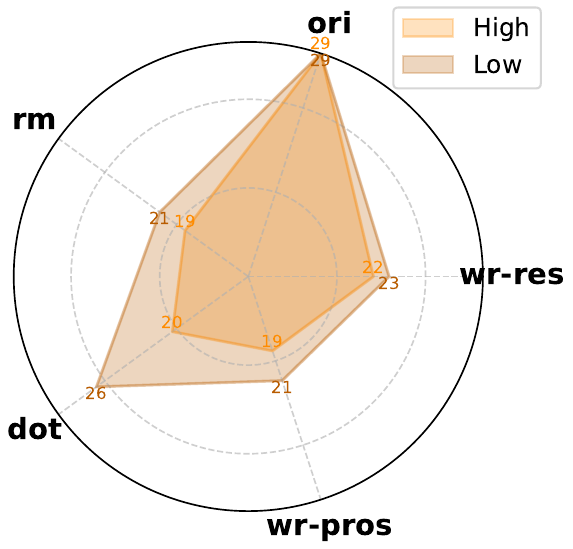}
           \caption{LLaMA2 on MathQA}
       \end{subfigure}%
       \hfill
       \begin{subfigure}[b]{0.24\textwidth}
           \centering
           \includegraphics[width=\textwidth]{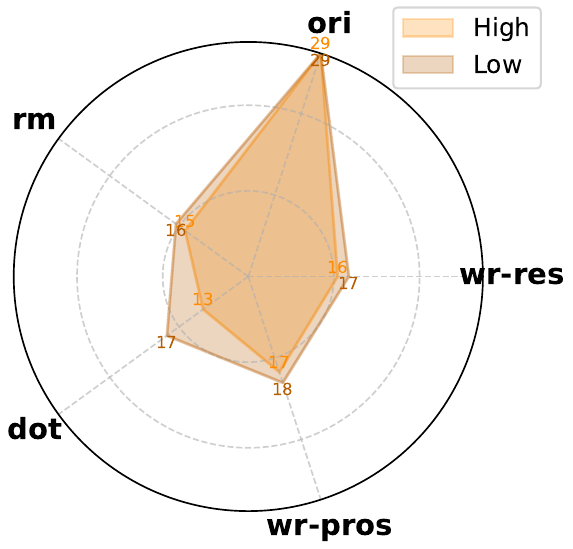}
           \caption{LLaMA2 on AQUA}
       \end{subfigure}%
       \caption{Accuracy of demonstrations for low and high CoSP-0 expressions after four types of modifications in the test set across different models and demos. (a) LLaMA2 on MathQA, (b) LLaMA on AQUA. }
       \label{fig:aba_mathqa_aqua}
       \vspace{-0.5cm}
   \end{figure}

Fig~\ref{fig:aba_mathqa_aqua} shows the results of LLaMA2 conducted on MathQA and AQUA, with the same setups as Sec~\ref{sec:exp4}.

Fig.~\ref{fig:aba_mathqa_aqua} indicates that the accuracy curve for low CoSP-0 expressions encompasses that for high CoSP-0 expressions in almost all scenarios, highlighting that alterations on low CoSP-0 expressions yield overall better performance outcomes compared to alterations on high CoSP-0 expressions.

\newpage
\onecolumn
\section{Selected Demonstrations}
\label{sec:app2}
This section presents the selected demonstrations in Sec~\ref{sec:exp4}. Expressions with \colorbox{blockblue}{a light blue background} have the lowest CoSP, those with \colorbox{blockorange}{an orange background} have the highest CoSP, and the remaining expressions are shown with \colorbox{blockgreen}{a light green background}.

\begin{tcolorbox}[
    enhanced,
    breakable,
    width=\textwidth,
    arc=3mm,
    boxrule=1pt,
    left=2mm,
    right=2mm,
    top=1mm,
    bottom=1mm,
    fontupper=\ttfamily,
    before skip=5mm,
    after skip=5mm,
    title=demo1, 
    coltitle=white, 
    colbacktitle=black, 
    fonttitle=\bfseries 
    ]
Question: \\
Sharon wants to get kitchen supplies. She admired Angela's kitchen supplies which consist of:
20 pots, 6 more than three times as many plates as the pots, and half as many cutlery as the plates. Sharon wants to buy: half as many pots as Angela, 20 less than three times as many plates as Angela, and twice as much cutlery as Angela. What is the total number of kitchen supplies Sharon wants to buy? \\
Answer:\\
Angela has \colorbox{blockblue}{6+3*20=<<6+3*20=66>>66 } plates. 
Angela has \colorbox{blockgreen}{1/2*66=<<1/2*66=33>>33} cutlery. 
Sharon wants to buy  \colorbox{blockgreen}{1/2*20=<<1/2*20=10>>10} pots.
Sharon wants to buy \colorbox{blockgreen}{3*66-20=<<3*66-20=178>>178} plates. 
Sharon wants to buy \colorbox{blockgreen}{2*33=<<2*33=66>>66} cutlery. 
Sharon wants to buy a total of \colorbox{blockorange}{10+178+66=<<10+178+66=254>>254} kitchen supplies.
\end{tcolorbox}

\begin{tcolorbox}[
    enhanced,
    breakable,
    width=\textwidth,
    arc=3mm,
    boxrule=1pt,
    left=2mm,
    right=2mm,
    top=1mm,
    bottom=1mm,
    fontupper=\ttfamily,
    before skip=5mm,
    after skip=5mm,
    title=demo2, 
    coltitle=white, 
    colbacktitle=black, 
    fonttitle=\bfseries 
    ]
    Question: \\
    Brittany, Alex, and Jamy all share 600 marbles divided between them in the ratio 3:5:7. If Brittany gives Alex half of her marbles, what's the total number of marbles that Alex has?\\
    Answer:\\
    The total ratio representing the number of marbles is \colorbox{blockblue}{3+5 +7 = <<3+5+7=15>>15}. From the ratio, the fraction representing the number of marbles that Brittany has is \colorbox{blockgreen}{3/15}, which is equal to \colorbox{blockgreen}{3/15*600 = <<3/15*600=120>>120} marbles.Alex has \colorbox{blockgreen}{5/15*600 = <<5/15*600=200>>200} marbles.If Brittany gives half of her marbles to Alex, Alex receives \colorbox{blockgreen}{1/2*120 = 60} marbles.After receiving 60 marbles from Brittany, Alex has \colorbox{blockorange}{200+60 = <<200+60=260>>260} marbles.
\end{tcolorbox}

\section{More Cases}
\label{sec:cases}
This section presents more demonstrations, with mathematical expressions with different CoSP shaded in different colors. The shading rule is the same as Appendix~\ref{sec:app2}, where the expressions with highest, medium, and lowest CoSP are shaded in \colorbox{blockorange}{orange}, \colorbox{blockgreen}{light green}, and \colorbox{blockblue}{light blue}.

Additionally, we demonstrate that there is minimal bias in CoSP and the positioning of expressions. Intuitively, expressions that are closer to the final answer tend to be more important, as they directly guide the model toward a specific output. In contrast, earlier expressions often represent intermediate or preliminary steps, which may have less influence on the final outcome. So the contribution of expressions may have bias with their positions. However, the CoSP of the expression shows low correlation with its position in the demonstration in most cases, indicating the feasibility of our framework.

\begin{tcolorbox}[
    enhanced,
    breakable,
    width=\textwidth,
    arc=3mm,
    boxrule=1pt,
    left=2mm,
    right=2mm,
    top=1mm,
    bottom=1mm,
    fontupper=\ttfamily,
    before skip=5mm,
    after skip=5mm,
    title=demo3, 
    coltitle=white, 
    colbacktitle=black, 
    fonttitle=\bfseries 
    ]
Question:\\
Sasha added 48 cards into a box. Her sister, Karen, then took out 1/6 of the cards Sasha added. If there are now 83 cards in the box, how many cards were originally in the box?\\
Answer: \\
Karen took out \colorbox{blockblue}{48/6 = <<48/6=8>>8} cards from the box.\\
Originally, the box had \colorbox{blockorange}{83-40 = <<83-40=43>>43} cards.
\end{tcolorbox}

\begin{tcolorbox}[
    enhanced,
    breakable,
    width=\textwidth,
    arc=3mm,
    boxrule=1pt,
    left=2mm,
    right=2mm,
    top=1mm,
    bottom=1mm,
    fontupper=\ttfamily,
    before skip=5mm,
    after skip=5mm,
    title=demo4, 
    coltitle=white, 
    colbacktitle=black, 
    fonttitle=\bfseries 
    ]
Question: \\
Coleen loved sprinkles. At the beginning of the day, she had twelve cans of sprinkles. After applying sprinkles to her hair, her clothing and her pets, she had 3 less than half as many cans of sprinkles as she started out with. How many cans of sprinkles remained?\\
Answer: \\
Half of twelve cans of sprinkles is \colorbox{blockorange}{12/2=<<12/2=6>>6}cans.\\
Three less than half as many cans of sprinkles is \colorbox{blockblue}{6-3=<<6-3=3>>3} cans of sprinkles.
\end{tcolorbox}

\begin{tcolorbox}[
    enhanced,
    breakable,
    width=\textwidth,
    arc=3mm,
    boxrule=1pt,
    left=2mm,
    right=2mm,
    top=1mm,
    bottom=1mm,
    fontupper=\ttfamily,
    before skip=5mm,
    after skip=5mm,
    title=demo5, 
    coltitle=white, 
    colbacktitle=black, 
    fonttitle=\bfseries 
    ]
Question: \\
Ali is collecting bottle caps. He has 125 bottle caps. He has red ones and green ones. If he has 50 red caps, what percentage of caps are green?\\
Answer:\\
He has 75 green caps because \colorbox{blockorange}{125 - 50 = <<125-50=75>>75}\\
The proportion of caps that are green is .6 because \colorbox{blockblue}{75 / 125 = <<75/125=.6>>.6}\\
The percentage that are green is 60 because \colorbox{blockgreen}{.6 x 100\% = <<60=60>>60\%}
\end{tcolorbox}

\begin{tcolorbox}[
    enhanced,
    breakable,
    width=\textwidth,
    arc=3mm,
    boxrule=1pt,
    left=2mm,
    right=2mm,
    top=1mm,
    bottom=1mm,
    fontupper=\ttfamily,
    before skip=5mm,
    after skip=5mm,
    title=demo6, 
    coltitle=white, 
    colbacktitle=black, 
    fonttitle=\bfseries 
    ]
Question: \\
Nathan plays amateur baseball. He played for 3 hours for two weeks, every day. His friend Tobias played for 5 hours every day, but only for one week. How many hours did Nathan and Tobias play in total?\\
Answer: \\
Two weeks are 14 days, so Nathan played for \colorbox{blockblue}{3 * 14 = <<14*3=42>>4} hours.\\
Tobias played for 7 days, so he played a total of \colorbox{blockorange}{5 * 7 = <<5*7=35>>35} hours.\\
Nathan and Tobias played together for \colorbox{blockgreen}{42 + 35 = <<42+35=77>>77} hours.
\end{tcolorbox}

\begin{tcolorbox}[
    enhanced,
    breakable,
    width=\textwidth,
    arc=3mm,
    boxrule=1pt,
    left=2mm,
    right=2mm,
    top=1mm,
    bottom=1mm,
    fontupper=\ttfamily,
    before skip=5mm,
    after skip=5mm,
    title=demo7, 
    coltitle=white, 
    colbacktitle=black, 
    fonttitle=\bfseries 
    ]
Question:\\
While bird watching, Gabrielle saw 5 robins, 4 cardinals, and 3 blue jays. Chase saw 2 robins, 3 blue jays, and 5 cardinals. How many more birds, in percentage, did Gabrielle saw than Chase?\\
Answer:\\
Gabrielle saw \colorbox{blockgreen}{5 + 4 + 3 = <<5+4+3=12>>12} birds.\\
Chase saw \colorbox{blockorange}{2 + 3 + 5 = <<2+3+5=10>>10} birds.\\
So, Gabrielle saw \colorbox{blockgreen}{12 - 10 = <<12-10=2>>2} more birds than Chase.\\
Therefore, Gabrielle saw \colorbox{blockblue}{2/10 x 100\% = 20\%} more birds than Chase.
\end{tcolorbox}

\begin{tcolorbox}[
    enhanced,
    breakable,
    width=\textwidth,
    arc=3mm,
    boxrule=1pt,
    left=2mm,
    right=2mm,
    top=1mm,
    bottom=1mm,
    fontupper=\ttfamily,
    before skip=5mm,
    after skip=5mm,
    title=demo8, 
    coltitle=white, 
    colbacktitle=black, 
    fonttitle=\bfseries 
    ]
 Question: \\
 Two alien spacecraft on a sightseeing tour of Earth left New Orleans airport at 3:00 pm to travel the 448-mile distance to Dallas by air. Traveling nonstop, the first spacecraft landed in Dallas at 3:30 pm, while the second spacecraft landed in Dallas thirty minutes later. Assuming both spacecraft traveled at constant speed, what was the difference in speed, in miles per hour, between the two spacecraft?\\
 Answer: \\
 The first spacecraft flew for 30 minutes, or \colorbox{blockgreen}{30/60=1/2 hour}.\\
 The second spacecraft flew for \colorbox{blockgreen}{30+30=<<30+30=60>>60} minutes, or 1 hour.\\
 Thus the first spacecraft traveled at a speed of 448 miles in 1/2 hour, or \colorbox{blockorange}{448/(1/2)=896} miles per hour.\\
 The second spacecraft traveled 448 miles in 1 hour, or \colorbox{blockblue}{448/1=<<448/1=448>>448} miles per hour.\\
 The difference in speed, in miles per hour, between the two spacecraft was \colorbox{blockgreen}{896-448=<<896-448=448>>448} miles per hour.
\end{tcolorbox}

\begin{tcolorbox}[
    enhanced,
    breakable,
    width=\textwidth,
    arc=3mm,
    boxrule=1pt,
    left=2mm,
    right=2mm,
    top=1mm,
    bottom=1mm,
    fontupper=\ttfamily,
    before skip=5mm,
    after skip=5mm,
    title=demo9, 
    coltitle=white, 
    colbacktitle=black, 
    fonttitle=\bfseries 
    ]
Question: \\
Julio has four bottles of orange soda and seven bottles of grape soda in his fridge. His friend Mateo has a bottle of orange soda and 3 bottles of grape soda in his fridge. If the amount of beverage in each bottle is 2 liters, how many more liters of soda does Julio have?\\
Answer: \\
Julio has \colorbox{blockorange}{4 * 2 = <<4*2=8>>8} liters of orange soda\\
Julio also has \colorbox{blockblue}{7 * 2 = <<7*2=14>>14} liters of grape soda.\\
Julio therefore has a total of \colorbox{blockgreen}{8 + 14 = <<8+14=22>>22} liters of soda\\
The amount of orange soda that Mateo has is \colorbox{blockgreen}{1 * 2 = <<1*2=2>>2} liters of orange soda\\
In addition, Mateo has \colorbox{blockgreen}{3 * 2 = <<3*2=6>>6} liters of grape soda.\\
In total, Mateo has \colorbox{blockgreen}{2 + 6 = <<2+6=8>>8} liters of soda.\\
This means that Julio has \colorbox{blockgreen}{22 - 8 = <<22-8=14>>14} liters more of soda
\end{tcolorbox}

\begin{tcolorbox}[
    enhanced,
    breakable,
    width=\textwidth,
    arc=3mm,
    boxrule=1pt,
    left=2mm,
    right=2mm,
    top=1mm,
    bottom=1mm,
    fontupper=\ttfamily,
    before skip=5mm,
    after skip=5mm,
    title=demo10, 
    coltitle=white, 
    colbacktitle=black, 
    fonttitle=\bfseries 
    ]
Question: \\
In a class of 30 students, the teacher polls the students on their favorite subject. 1/5 of the students like Math, and 1/3 like English. 1/7 of the remaining students like Science. The rest don't have a favorite subject. How many students don't have a favorite subject?\\
Answer: \\
\colorbox{blockgreen}{30 x 1/5 = <<30*1/5=6>>6} students like Math.\\
\colorbox{blockgreen}{30 x 1/3 = <<30*1/3=10>>10} students like English.\\
So, \colorbox{blockgreen}{6 + 10 = <<6+10=16>>16} students like either Math or English.\\
Thus, \colorbox{blockgreen}{30 - 16 = <<30-16=14>>14} students neither like Math nor English.\\
Since \colorbox{blockblue}{1/7} of the remaining like Science, therefore \colorbox{blockgreen}{14 x 1/7 = <<14*1/7=2>>2} students like Science.\\
Hence, \colorbox{blockorange}{14 - 2 = <<14-2=12>>12} students neither likes the 3 subjects.
\end{tcolorbox}

\newpage
\section{Qualitative Analysis Examples}
\label{sec:quality_cases}
This section presents Example 1 and Example 2 used in Sec~\ref{sec:quality}.
\begin{tcolorbox}[
    enhanced,
    breakable,
    width=\textwidth,
    arc=3mm,
    boxrule=1pt,
    left=2mm,
    right=2mm,
    top=1mm,
    bottom=1mm,
    fontupper=\ttfamily,
    before skip=5mm,
    after skip=5mm,
    title=Example1, 
    coltitle=white, 
    colbacktitle=black, 
    fonttitle=\bfseries 
    ]
    Question: \\
    a, b, k start from the same place and travel in the same direction at speeds of 30 km / hr, 40 km / hr, 60 km / hr respectively. b starts three hours after a. if b and k overtake a at the same instant, how many hours after a did k start? a ) 3 , b ) 4.5 , c ) 6 , d ) d ) 5.5 , e ) e ) 5 \\
    Answer: \\
    "the table you made doesn't make sense to me. all three meet at the same point means the distance they cover is the same. we know their rates are 30, 40 and 60. say the time taken by b is t hrs. then a takes 3 + t hrs. and we need to find the time taken by k. distance covered by a = distance covered by b \colorbox{blockorange}{30 * ( 3 + t ) = 40 * t t = 9} hrs distance covered by b = distance covered by \colorbox{blockorange}{k 40 * t = 60} * time taken by k time taken by \colorbox{blockorange}{k = 40 * 9 / 60 = 6} hrs time taken by \colorbox{blockorange}{a = 3 + t = 3 + 9 = 12} hrs time taken by k = 6 hrs so k starts \colorbox{blockorange}{12 - 6 = 6} hrs after a . ( answer c )" 
\end{tcolorbox}

\begin{tcolorbox}[
    enhanced,
    breakable,
    width=\textwidth,
    arc=3mm,
    boxrule=1pt,
    left=2mm,
    right=2mm,
    top=1mm,
    bottom=1mm,
    fontupper=\ttfamily,
    before skip=5mm,
    after skip=5mm,
    title=Example2, 
    coltitle=white, 
    colbacktitle=black, 
    fonttitle=\bfseries 
    ]
    Question: \\
    Question: of 70 players on a football team, 46 are throwers. the rest of the team is divided so one third are left - handed and the rest are right handed. assuming that all throwers are right handed, how many right - handed players are there total? a ) 54 , b ) 59 , c ) 63 , d ) 71 , e ) 62 \\
    Answer:\\
    "total = 70 thrower = 46 rest = \colorbox{blockblue}{70 - 46 = 24} left handed \colorbox{blockblue}{= 24 / 3 = 8} right handed = 16 if all thrower are right handed then total right handed is \colorbox{blockblue}{46 + 16 = 62} so e. 62 is the right answer"
\end{tcolorbox}

\end{document}